\documentclass[10pt,journal,compsoc]{IEEEtran}
\ifCLASSOPTIONcompsoc
\usepackage[nocompress]{cite}
\else
\usepackage{cite}
\fi
\ifCLASSINFOpdf
\else
\fi

\usepackage{booktabs,multirow}
\usepackage{graphicx}  
\usepackage{array}

\usepackage{utfsym}
\usepackage{color,colortbl}
\definecolor{blush}{rgb}{0.87, 0.36, 0.51}
\usepackage[colorlinks,citecolor=green]{hyperref}

\usepackage{amsmath}
\usepackage{amssymb}
\definecolor{bblue}{rgb}{0.36, 0.51, 0.87}
\definecolor{ggray}{rgb}{0.88, 0.87, 0.87}
\newcolumntype{a}{>{\columncolor{ggray}}c}

\usepackage[capitalize]{cleveref}
\crefname{section}{Sec.}{Secs.}
\Crefname{section}{Section}{Sections}
\Crefname{table}{Table}{Tables}
\crefname{table}{Tab.}{Tabs.}

\usepackage{bbm}

\usepackage{forest}

\usepackage{amsthm}
\usepackage{cancel}
\theoremstyle{definition}



\newcommand{\bs}[1]{\ensuremath{\boldsymbol{#1}}}
\newcommand{\promptFeat}{\mathsf{TrgEmb}}

\newcommand{\mc}[1]{\ensuremath{\mathcal{#1}}}

\newcommand{\mb}[1]{\ensuremath{\mathbf{#1}}}

\makeatletter
\newcommand*\bigcdot{\mathpalette\bigcdot@{.5}}
\newcommand*\bigcdot@[2]{\mathbin{\vcenter{\hbox{\scalebox{#2}{$\m@th#1\bullet$}}}}}
\makeatother

\usepackage{makecell}
\usepackage{ragged2e}
\usepackage{subfigure}

\begin{document}
\title{Advances in Multimodal Adaptation and Generalization: From Traditional Approaches \\to Foundation Models}
% Mitigating Domain Shifts with Multimodal Learning Methods: A Survey
% Advances in Multimodal Adaptation and Generalization: From Traditional Approaches to Foundation Models
% From Traditional Methods to Foundation Models: A Survey on Multimodal Domain Adaptation and Generalization

\author{
Hao Dong, Moru Liu, Kaiyang Zhou, Eleni Chatzi, Juho Kannala, Cyrill Stachniss and Olga Fink
% ,~\IEEEmembership{Senior Member,~IEEE,}
\IEEEcompsocitemizethanks{
\IEEEcompsocthanksitem Hao Dong and Eleni Chatzi are with ETH Z\"urich, Switzerland.% (e-mail: hao.dong@ibk.baug.ethz.ch; chatzi@ibk.baug.ethz.ch). 
\IEEEcompsocthanksitem Moru Liu is with the Technical University of Munich, Germany.% (e-mail:  moru.liu@tum.de).
\IEEEcompsocthanksitem Kaiyang Zhou is with Hong Kong Baptist University, China.
\IEEEcompsocthanksitem Juho Kannala is with Aalto University and University of Oulu, Finland.
\IEEEcompsocthanksitem Cyrill Stachniss is with the University of Bonn, Germany, the University of Oxford, UK, and the Lamarr Institute for Machine Learning and Artificial Intelligence, Germany.
\IEEEcompsocthanksitem Olga Fink is with EPFL, Switzerland.% (e-mail: olga.fink@epfl.ch).
}
}

% \markboth{}
% {Dong \MakeLowercase{\textit{et al.}}: A Comprehensive Survey on }

\IEEEtitleabstractindextext{%
\begin{abstract}
\justifying
Domain adaptation and generalization are crucial for real-world applications, such as autonomous driving and medical imaging where the model must operate reliably across environments with distinct data distributions. However, these tasks are challenging because the model needs to overcome various domain gaps caused by variations in, for example, lighting, weather, sensor configurations, and so on. Addressing domain gaps simultaneously in different modalities, known as multimodal domain adaptation and generalization, is even more challenging due to unique challenges in different modalities. Over the past few years, significant progress has been made in these areas, with applications ranging from action recognition to semantic segmentation, and more. Recently, the emergence of large-scale pre-trained multimodal foundation models, such as CLIP, has inspired numerous research studies, which leverage these models to enhance downstream adaptation and generalization. This survey summarizes recent advances in multimodal adaptation and generalization, particularly how these areas evolve from traditional approaches to foundation models. Specifically, this survey covers (1) multimodal domain adaptation, (2) multimodal test-time adaptation, (3) multimodal domain generalization, (4) domain adaptation and generalization with the help of multimodal foundation models, and (5) adaptation of multimodal foundation models. For each topic, we formally define the problem and give a thorough review of existing methods. Additionally, we analyze relevant datasets and applications, highlighting open challenges and potential future research directions. We also maintain an active repository that contains up-to-date literature and supports research activities in these fields at \url{https://github.com/donghao51/Awesome-Multimodal-Adaptation}.
\end{abstract}

% Note that keywords are not normally used for peerreview papers.
}

\maketitle\IEEEdisplaynontitleabstractindextext
\IEEEpeerreviewmaketitle

\ifCLASSOPTIONcompsoc
\IEEEraisesectionheading{\section{Introduction}\label{sec:introduction}}
\else
\section{Introduction}
\label{sec:introduction}
\fi

\begin{figure}[t]
  \centering
  \includegraphics[width=\linewidth]{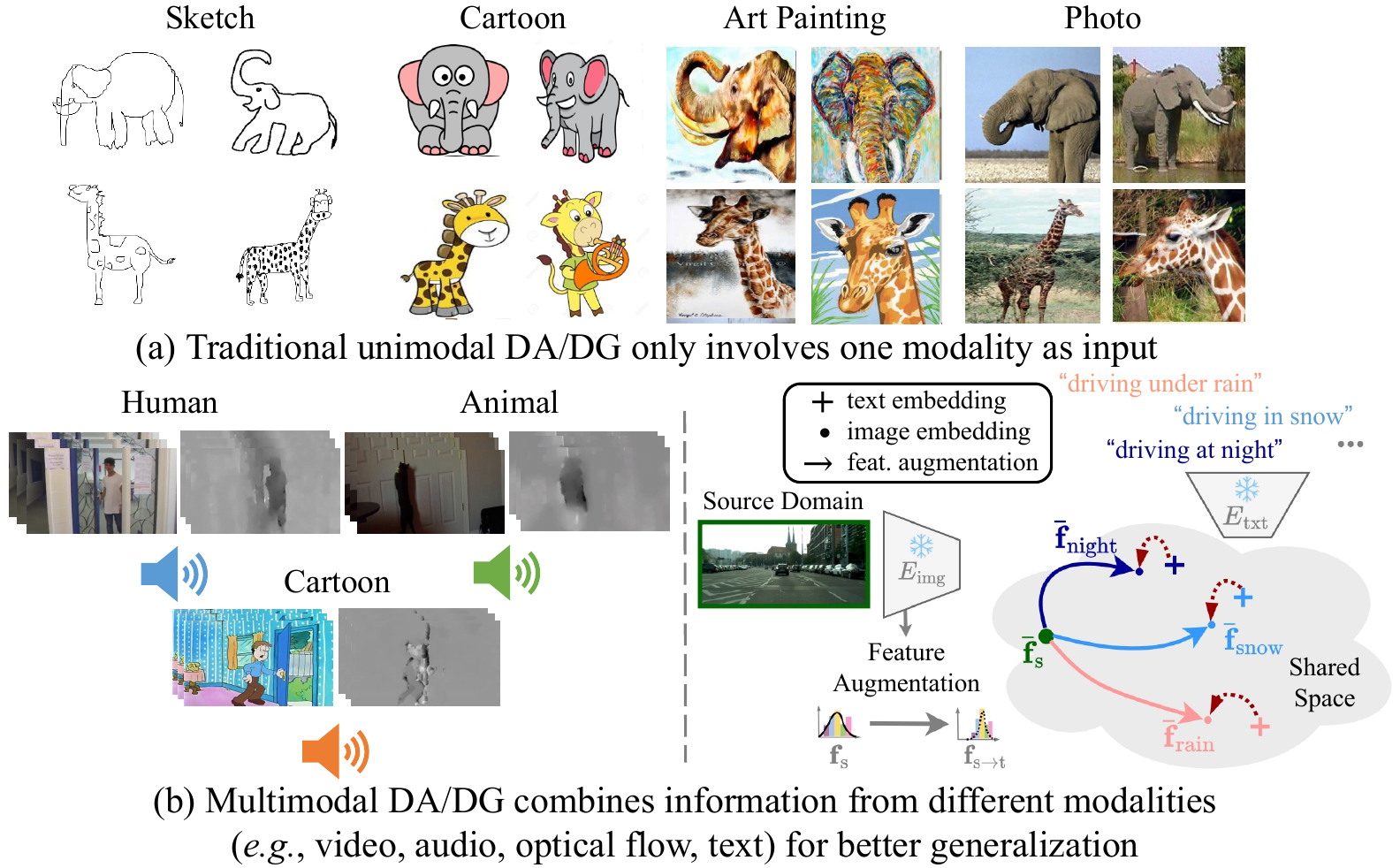}
   \vspace{-0.8cm}
   \caption{The difference between unimodal and multimodal DA/DG.}
   \label{fig:moti}
\end{figure}

\IEEEPARstart{D}{omain} adaptation (DA) and domain generalization (DG) have attracted significant attention in the research community~\cite{wang2018deep,wang2022generalizing}. In real-world applications such as robotics~\cite{dong2022dr,dong2023jras}, action recognition~\cite{Damen2018EPICKITCHENS}, and anomaly detection~\cite{sun2024ano,dong2023nngmix}, it is essential for models trained on limited source domains to perform well on novel target domains. To address distribution shift challenges, numerous DA and DG algorithms have been proposed, including distribution alignment~\cite{long2015learning}, domain-invariant feature learning~\cite{pmlr-v28-muandet13}, feature disentanglement~\cite{piratla2020efficient,zunino2021explainable}, data augmentation~\cite{zhang2018mixup,wang2021learning}, and meta-learning~\cite{li2018learning}. Most of these algorithms are designed for unimodal data, such as images or time series data. However, the real world is inherently multimodal and it is necessary to address multimodal domain adaptation (MMDA) and generalization (MMDG) across multiple modalities, including audio-video~\cite{Kazakos_2019_ICCV}, image-language~\cite{radford2021learning}, and LiDAR-camera~\cite{SuperFusion}. \cref{fig:moti} illustrates the distinction between unimodal and multimodal DA/DG, where MMDA and MMDG integrate information from multiple modalities to enhance generalization ability. {Multimodal DA/DG introduces unique complexities that cannot be viewed as simple extensions of unimodal problems. These include: (i) heterogeneous domain shifts, where different modalities (e.g., video and audio) are affected by distinct types and degrees of distribution shift; (ii) the difficulty of preserving cross-modal dependencies during alignment, since naive alignment of each modality in isolation can disrupt the critical relationships between them; and (iii) practical challenges such as modality imbalance and missingness, which are specific to multimodal settings. These challenges highlight that multimodal DA/DG is not merely a variant of traditional unimodal DA/DG, but a field with its own distinct set of challenges, warranting a dedicated and in-depth survey.}
%MMDA/DG is more challenging than unimodal DA/DG due to the need to align heterogeneous modalities with distinct domain shifts while preserving cross-modal dependencies. The presence of missing or noisy modalities and the increased computational complexity of simultaneous cross-domain and cross-modal alignment further complicate adaptation.

\begin{figure*}[t]
  \centering
  \includegraphics[width=\linewidth]{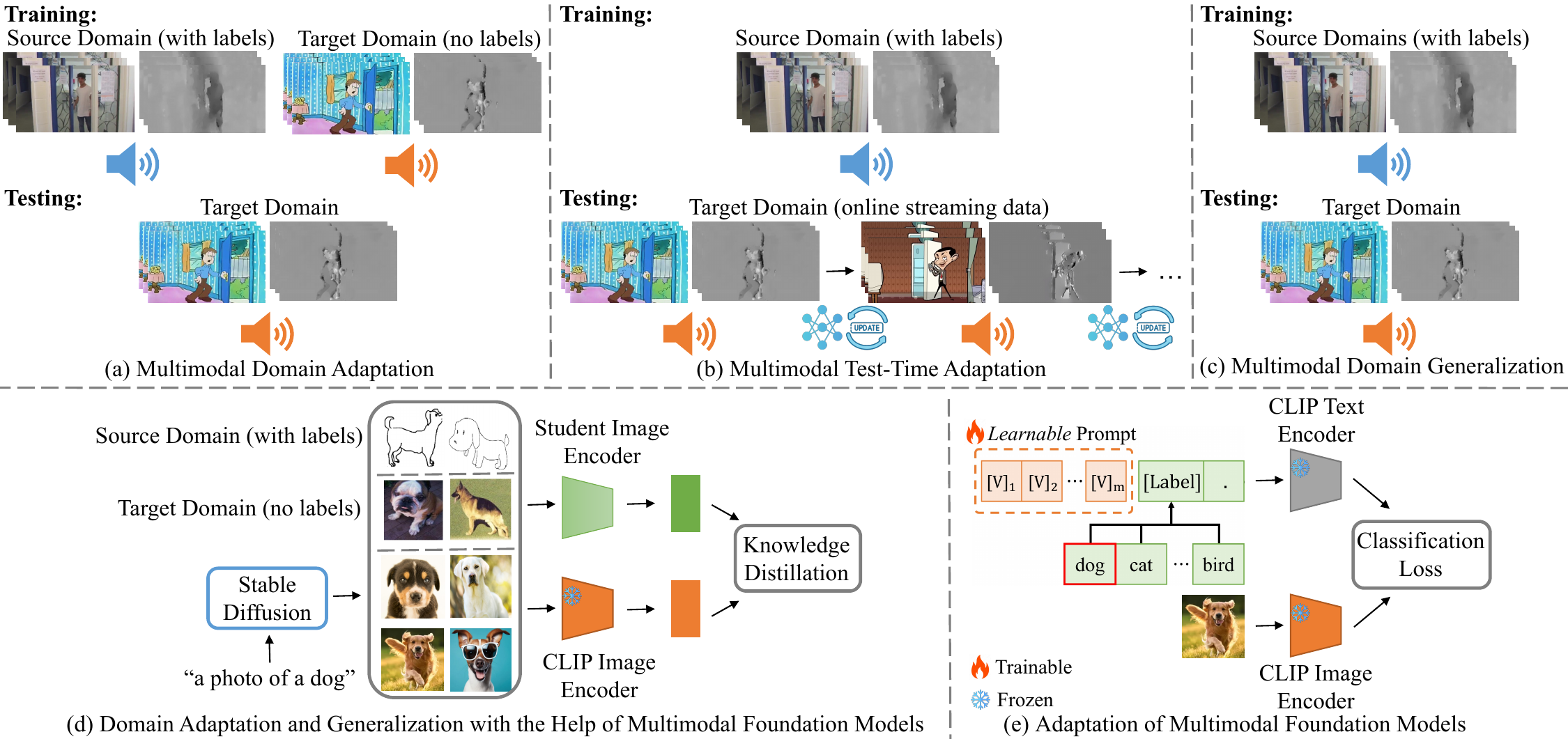}
   \vspace{-0.8cm}
   \caption{Five multimodal adaptation scenarios are discussed in this survey. (a) Multimodal domain adaptation, (b) Multimodal test-time adaptation, and (c) Multimodal domain generalization, which represent traditional multimodal settings with varying access to source and target domain data. Additionally, we examine more recent foundation models including (d) unimodal domain adaptation and generalization assisted by multimodal foundation models, and (e) the adaptation of multimodal foundation models to downstream tasks. }
   \label{fig:frame}
\end{figure*}

In recent years, MMDA and MMDG have achieved significant progress in areas such as action recognition~\cite{munro2020multi}, semantic segmentation~\cite{jaritz2020xmuda}, {cross-modal retrieval~\cite{liu2021adaptive}, and generative modeling~\cite{zhao2025doracycle}}. %A central challenge in MMDA and MMDG is effectively leveraging complementary information from diverse modalities to enhance generalization performance—an area where unimodal DA and DG approaches often fall short.  %For instance, the approach by Munro and Damen~\cite{munro2020multi} incorporates the within-modal adversarial alignment alongside multimodal self-supervised alignment for MMDA. %Multimodal test-time adaptation (MMTTA)~\cite{yang2023test} is a special form of MMDA that focuses on adapting a pre-trained source multimodal model to a target domain online, without accessing source domain data.
%Similarly, MOOSA~\cite{dong2024towards} introduces multimodal pretext tasks, such as Masked Cross-modal Translation and Multimodal Jigsaw Puzzles to improve generalization and open-set detection for MMDG. 
Besides, the emergence of large-scale multimodal foundation models (MFMs), such as contrastive language–image pretraining (CLIP)~\cite{radford2021learning} and stable diffusion~\cite{rombach2022high}, has opened new research directions for DA and DG. These efforts aim to enhance generalization capabilities using MFMs or adapt MFMs to downstream tasks. %, also referred to as MMDA or MMDG in this paper. 
For example, Dunlap et al.~\cite{dunlap2023using} extend image embeddings to unseen domains using language, while Huang et al.~\cite{huang2023sentence} distill CLIP’s knowledge into a smaller student model for domain generalization. Additionally, Zhou et al.~\cite{zhou2022learning} adapt CLIP-like vision-language models (VLMs) for downstream image recognition by modeling a prompt’s context words with learnable vectors.

\begin{figure*}[t!]
	\centering
	\resizebox{\textwidth}{!}{
	\begin{forest}
  for tree={
  grow=east,
  reversed=true,
  anchor=base west,
  parent anchor=east,
  child anchor=west,
  base=left,
  font=\small,
  rectangle,
  draw,
  rounded corners,align=left,
  minimum width=2.5em,
  inner xsep=4pt,
  inner ysep=1pt
  },
  where level=1{fill=blue!10}{},
  where level=2{font=\footnotesize,fill=pink!30}{},
  where level=3{font=\footnotesize,yshift=0.26pt,fill=yellow!20}{},
  where level=4{font=\footnotesize,yshift=0.26pt,fill=gray!10}{},
  [Multimodal Adaptation\\and Generalization,fill=blue!20
[Multimodal\\Domain\\Adaptation,fill=green!20
    [Action\\Recognition
        [Domain-Adversarial Learning
            [e.g. MM-SADA~\cite{munro2020multi}/MDANN~\cite{U4DA}/PMC~\cite{9334409}/MD-DMD~\cite{10.1145/3503161.3548313}
            ]
        ]
        [Contrastive Learning
          [e.g. STCDA~\cite{9577559}/Kim et al.~\cite{kim2021learning}
          ]
        ]
        [Cross-Modal  Interaction
            [e.g. MOOSA~\cite{dong2024towards}/DLMM~\cite{10.1145/3474085.3475660}/MTRAN~\cite{10.1145/3503161.3548009}/Zhang et al.~\cite{zhang2022audio}
            ]
        ]
    ]
    [Semantic\\Segmentation
        [xMUDA and\\Its Extensions
            [e.g. xMUDA~\cite{jaritz2020xmuda}/Dual-cross~\cite{10.1145/3503161.3547990}/CoMoDaL~\cite{10.1145/3581783.3612320}/MoPA~\cite{cao2024mopa}/\\MM2D3D~\cite{cardace2023exploiting}/SUMMIT~\cite{10377604}/SSE-xMUDA~\cite{10.1145/3503161.3547987}/CMCL~\cite{xing2023cross}/Mx2M~\cite{zhang2023mx2m}
            ]
        ]
        [Domain-Adversarial Learning
         [e.g. DsCML~\cite{peng2021sparse}/AUDA~\cite{LIU2021211}/DualCross ~\cite{man2023dualcross} 
         ]
        ]
        [Cross-Modal  Interaction
            [e.g. Drive\&Segment~\cite{vobecky2022drive}/CrossMatch~\cite{10376488}/MISFIT~\cite{rizzoli2024source}
            ]
        ]
    ]
    [Other Tasks
            [e.g. MMAN~\cite{ma2019deep}/OSAN~\cite{10204066}/Amanda~\cite{zhang-etal-2024-amanda},fill=gray!10 
            ]
    ]
]
[Multimodal Test-time\\Adaptation,fill=green!20
        [Action
Recognition
            [e.g. READ~\cite{yang2023test}/AEO~\cite{dong2025aeo}/MC-TTA~\cite{xiong2024modality}/2LTTA~\cite{lei2024twolevel}/ MiDl~\cite{anonymous2025testtime},fill=gray!10 
            ]
    ]
        [Semantic Segmentation\\and Other Tasks
            [e.g. MM-TTA~\cite{shin2022mm}/CoMAC~\cite{cao2023multi}/Latte~\cite{cao2024reliable}/ProxyTTA~\cite{park2024test}/HTT~\cite{wang2024heterogeneous},fill=gray!10 
            ]
    ]
]
[Multimodal Domain\\Generalization,fill=green!20
        [Action
Recognition
            [e.g. SimMMDG~\cite{dong2023simmmdg}/MOOSA~\cite{dong2024towards}/RNA-Net~\cite{planamente2022domain}/CMRF~\cite{fan2024crossmodal},fill=gray!10 
            ]
    ]
        [Semantic Segmentation
            [e.g. BEV-DG~\cite{li2023bev},fill=gray!10 
            ]
    ]
]
[DA and DG\\with the Help\\of MFMs,fill=green!20
    [Data Augmentation
        [Input Space
            [e.g. DGInStyle~\cite{jia2025dginstyle}/CDGA~\cite{hemati2024leveraging}/CLOUDS~\cite{benigmim2024collaborating}/ODG-CLIP~\cite{singha2024unknown}
            ]
        ]
        [Feature Space
          [e.g. LADS~\cite{dunlap2023using}/PODA~\cite{fahes2023poda}/PromptStyler~\cite{cho2023promptstyler}/ULDA~\cite{yang2024unified}
          ]
        ]
    ]
    [Knowledge\\Distillation
        [Teacher-Student\\Model
            [e.g. DALL-V~\cite{zara2023unreasonable}/RISE~\cite{huang2023sentence}/SCI-PD~\cite{chen2024practicaldg}/DIFO~\cite{tang2024source}/VL2V-ADiP~\cite{addepalli2024leveraging}
            ]
        ]
    ]
    [Learning\\Strategies
        [Prompt-based
            [e.g. DPL~\cite{zhang2023domain}/DAPL~\cite{ge2023domain}/MPA~\cite{chen2023multi}/Cheng et al. \cite{cheng2024disentangled}/TPL~\cite{wang2024transitive}
            ]
        ]
        [Prior-based
         [e.g. CSI~\cite{lim2024cross}/VFMSeg~\cite{xu2024visual}/CLIP-Div~\cite{zhu2024clip}/Peng et al.~\cite{peng2025learning}/Diffusion-TTA~\cite{prabhudesai2023diffusion}
         ]
        ]
        [Refinement-based
            [e.g. Lai et al.~\cite{lai2024empowering}/ReCLIP~\cite{hu2024reclip}/MADM~\cite{xia2024unsupervised}/Co-learn++~\cite{zhang2024source}
            ]
        ]
        [Others
            [e.g. MIRO~\cite{cha2022domain}/CLIPood~\cite{shu2023clipood}/PADCLIP~\cite{lai2023padclip}/FormerStereo~\cite{zhang2025learning}/CLIPCEIL~\cite{yuclipceil}
            ]
        ]
    ]
]
[Adaptation\\of MFMs,fill=green!20
    [Prompt-based
        [Text Prompt
            [e.g. CoOp~\cite{zhou2022learning}/CoCoOp~\cite{zhou2022conditional}/SubPT~\cite{ma2022understanding}/LASP~\cite{bulat2022language}/ProDA~\cite{lu2022prompt}/BPT~\cite{derakhshani2022variational}
            ]
        ]
        [Visual Prompt
          [e.g. VP~\cite{bahng2022exploring}/RePrompt~\cite{rong2023retrieval}/EVP~\cite{wu2022unleashing}
          ]
        ]
        [Text-Visual Prompt
          [e.g. UPT~\cite{zang2022unified}/MVLPT~\cite{shen2022multitask}/MaPLE~\cite{khattak2022maple}/DPT~\cite{xing2023dual}/CasPL~\cite{wu2025cascade}
          ]
        ]
    ]
    [Adapter-based
        [Feature Adapter
            [e.g. Clip-Adapter~\cite{gao2021clip}/SVL-Adapter~\cite{pantazis2022svl}/CLIPPR~\cite{kahana2022improving}/SgVA-CLIP~\cite{peng2022sgva}
            ]
        ]
        [Training-free
            [e.g. TDA~\cite{karmanov2024efficient}/Tip-Adapter~\cite{zhang2021tip} 
            ]
        ]
    ]
    [Others
        [Fine-tuning
            [e.g. Wise-FT~\cite{wortsman2022robust}/MUST~\cite{li2022masked}
            ]
        ]
        [Training-free
            [e.g. SuS-X~\cite{udandarao2022sus}/CALIP~\cite{guo2022calip}/MTA~\cite{zanella2024test}/DMN~\cite{zhang2024dual2}/GDA~\cite{wang2024hard}
            ]
        ]
        [LLM-based
            [e.g. CuPL~\cite{pratt2022does}/VCD~\cite{menon2022visual}/REAL~\cite{parashar2024neglected}
            ]
        ]
        [TTA of VLMs
            [e.g. TPT~\cite{shutest}/WATT~\cite{osowiechi2024watt}/DiffTPT~\cite{feng2023diverse}/ZERO~\cite{farina2024frustratingly}/SwapPrompt~\cite{ma2024swapprompt} ]
        ]
        [Dense Prediction
            [e.g. MaskCLIP~\cite{zhou2022extract}/Zhang et al.~\cite{zhang2024improving}/DenseCLIP~\cite{rao2022denseclip}
            ]
        ]
    ]
    ]
]
\end{forest}

        
	}
   \vspace{-0.7cm}
	\caption{Taxonomy of multimodal adaptation and generalization methods, with five novel scenarios involved.}
	\label{fig-over}
\end{figure*}
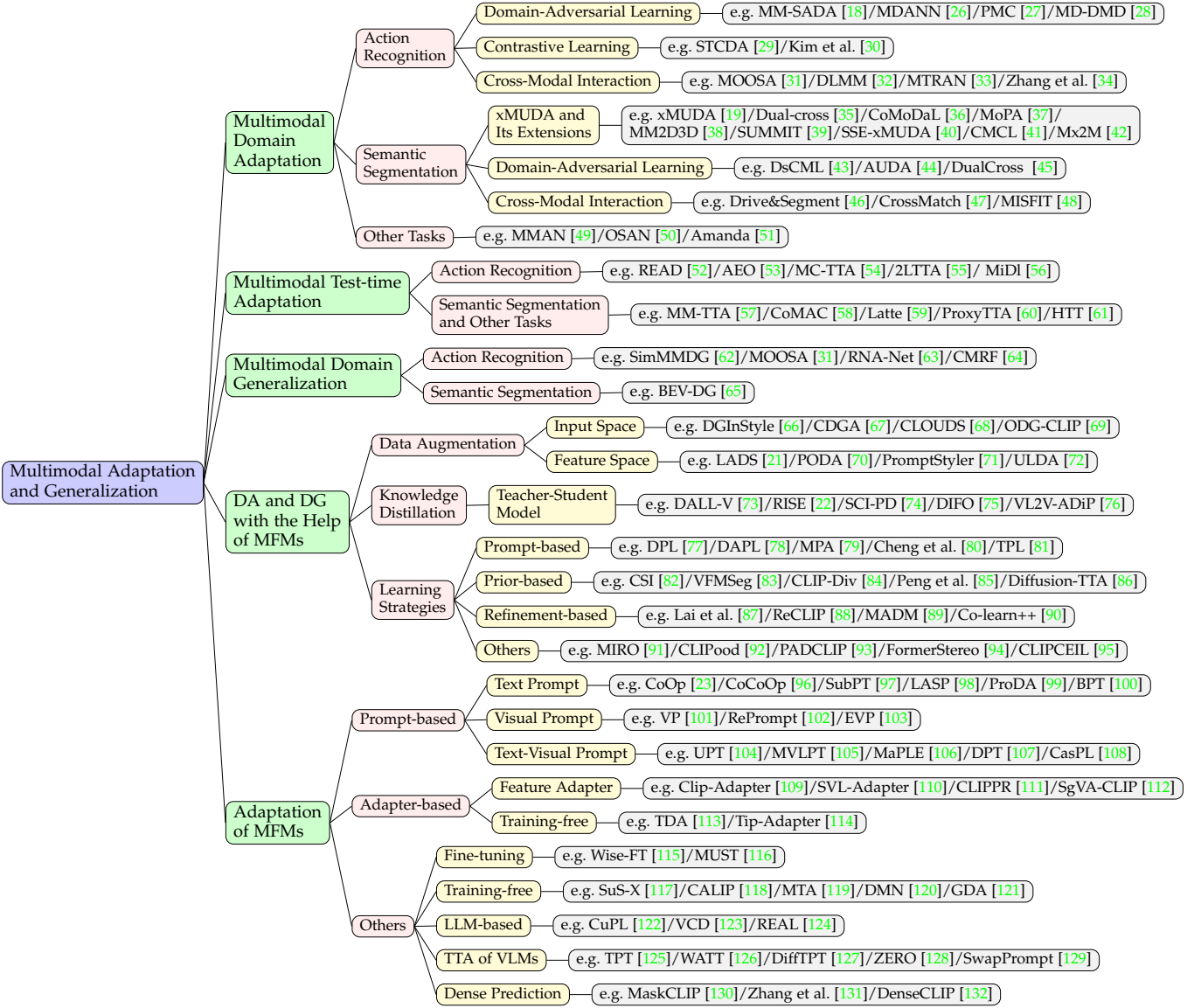

Despite significant progress made in the field recently, there is no comprehensive survey that summarizes the main ideas of multimodal adaptation and generalization. This survey paper aims to provide a detailed literature review of algorithms developed over the last decade and to offer insights into existing research gaps and future directions. 
This paper covers five adaptation scenarios (\cref{fig:frame} and \cref{fig-over}) and is organized as follows. \cref{sec:related} discusses related research areas, with a generic framework proposed in \cref{sec:GenericFramework}. \cref{sec:mmda} introduces the multimodal domain adaptation problem and highlights major solutions for action recognition and semantic segmentation. \cref{sec:mmtta} and \cref{sec:mmdg} present representative methods for multimodal test-time adaptation (MMTTA, a special form of MMDA that focuses on adapting a pre-trained source multimodal model to a target domain online, without accessing source domain data) and domain generalization, respectively. \cref{sec:dadgf} explores how multimodal foundation models can help improve DA and DG. \cref{sec:adaf} reviews methods for adapting MFMs to downstream tasks.
\cref{sec:Revisiting} revisits traditional DA/DG paradigms in the era of MFMs.
\cref{sec:application} summarizes major applications and datasets. 
Finally, we outline potential future directions in \cref{sec:future} and conclude the paper in \cref{sec:con}.

\noindent\textbf{Comparison with previous surveys.} While our survey contributes to the broader areas of DA and DG, which have been reviewed in prior works~\cite{wang2018deep,wang2022generalizing,zhou2022dg,dong2025adapting}, our specific focus is on multimodal adaptation and generalization, i.e. methods that involve multiple modalities. The recent survey by Zhang et al.~\cite{zhang2024vision} only covers an overview of the adaptation of VLMs before 2023. In contrast, we unify the discussion of traditional approaches for the novel MMDA, MMTTA, and MMDG setups, the role of advanced MFMs in enhancing DA and DG, as well as the recent methods for adapting MFMs to downstream tasks.

\section{Related Research Topics}
\label{sec:related}

\subsection{Domain Adaptation}  

%Domain Adaptation (DA) and Domain Generalization (DG) are two fundamental paradigms for addressing domain shifts between source and target domains. 
Domain adaptation seeks to enhance model performance in the target domain by leveraging labeled source data and unlabeled target data~\cite{wang2018deep}. Traditional DA methods typically focus on computer vision applications with images as the main input. Common approaches include aligning feature distributions using discrepancy metrics~\cite{kang2019contrastive}, employing adversarial learning in input or feature spaces~\cite{chen2020adversarial}, and utilizing reconstruction-based methods~\cite{yang2020label}. In addition, techniques such as data augmentation~\cite{zhang2018mixup} and self-training~\cite{sun2022safe}, have also been extensively explored. Depending on assumptions about label set relationships between the source and target domains, DA is further categorized into partial-set~\cite{cao2018partial}, open-set~\cite{nejjar2024sfosda}, and universal DA~\cite{saito2020universal}, which are more practical and challenging. %For further information, we refer the reader to~\cite{wang2018deep,li2024comprehensive}. 

\subsection{Domain Generalization}  
Domain generalization aims to generalize models to unseen target domains without accessing target data during training. DG methods can be broadly grouped into data manipulation, representation learning, and learning strategies~\cite{wang2022generalizing}. Data manipulation methods, such as~\cite{tobin2017domain}, enhance data diversity, while representation learning approaches~\cite{tzeng2014deep} focus on extracting domain-invariant features. Additionally, learning strategies like meta-learning~\cite{li2018learning} and self-supervised learning~\cite{carlucci2019domain} have demonstrated improved generalization performance across domains. Similar to DA, some works~\cite{shu2021open} also address the open-set DG problem where the target domain has private classes. %For further information on DG, we refer the reader to~\cite{zhou2022dg,wang2022generalizing}. Traditional DG methods also only focus on unimodal scenarios while this survey aims to give a comprehensive review of MMDA and MMDG methods. 

 %Closed-set DA assumes identical label sets for source and target domains, partial-set DA handles target domains with a subset of source labels, open-set DA considers additional unseen target labels, and universal DA adapts without prior assumptions about label set relationships.
 
% \textcolor{red}{[don't focus too much on MMDA and MMDG here]} While DA and DG have shown success in single-modal settings, their application to multimodal scenarios is increasingly relevant in real-world applications, such as autonomous driving, where LiDAR and camera data must be fused~\cite{SuperFusion}. However, the challenges in multimodal scenarios—such as modality inconsistencies, missing modalities, and complexities in cross-modal integration—require additional considerations. These challenges are particularly prominent in Multimodal Domain Adaptation (MMDA) and Multimodal Domain Generalization (MMDG), which aim to exploit cross-modal synergies to improve adaptability and generalization.

% Although multimodal settings are not the primary focus here, addressing the persistent sim-to-real gap and improving cross-modal alignment remain significant challenges for advancing robust DA and DG frameworks in both single- and multimodal settings.

\subsection{Test-time Adaptation}  
Test-time adaptation (TTA) seeks to adapt a pre-trained model on the source domain online, addressing distribution shifts without requiring access to either source data or target labels. Online TTA methods~\cite{wang2021tent,yuan2023robust} update specific model parameters using incoming test samples based on unsupervised objectives such as entropy minimization and pseudo-labels. Robust TTA methods~\cite{niu2022efficient,zhou2023ods} address more complex and practical scenarios, including label shifts, single-sample adaptation, and mixed domain shifts. Continual TTA approaches~\cite{wang2022continual,gan2023decorate} target the continual and evolving distribution shifts encountered over test time, which is particularly prevalent in real-world applications. %For further information, we refer the reader to the survey papers~\cite{liang2024comprehensive,wang2024search}.

%\subsection{Multimodal Learning}
%Multimodal learning leverages the complementary strengths of diverse modalities to enhance representation learning and contextual understanding. Prominent multimodal learning directions can be categorized into multimodal representation learning~\cite{peng2017ccl, bachman2019learning}, fusion methods~\cite{liu2018efficient, bai2022transfusion}, alignment~\cite{cirik2018visual, haresh2021learning}, etc. For further information, we refer the reader to~\cite{xu2023multimodal,liang2024foundations}. %These methods have also been extended to  MMDA~\cite{munro2020multi} and MMDG~\cite{dong2024towards} approaches. 

%Despite its potential, multimodal learning faces inherent challenges, including modality inconsistencies, missing modalities, inter-modal conflicts, and complexities in cross-modal integration, which remain critical directions for future research.

\subsection{Self-supervised Learning}
%Both unimodal and multimodal methods often depend heavily on extensive human annotations for effective training. 
Self-supervised learning (SSL) aims to learn from unlabeled data by obtaining supervision signals from pretext tasks, such as predicting transformations~\cite{doersch2015unsupervised, zhang2016colorful}, reconstructing missing components~\cite{li2021selfdoc, lu2022unified}, or optimizing contrastive objectives~\cite{tian2020contrastive, Yuan_2021_CVPR}. By capturing intrinsic data structures, SSL enables learning robust and domain-invariant representations, making it an essential component for DA and DG. In the multimodal context, SSL is exploited through tasks such as multimodal alignment~\cite{arandjelovic2017look}, cross-modal translation~\cite{alayrac2020self}, and relative norm alignment~\cite{owens2016ambient}. These pretext tasks have been recently effectively integrated into MMDA and MMDG frameworks, including methods such as MOOSA~\cite{dong2024towards} and MM-SADA~\cite{munro2020multi}. For further information on SSL, we refer the reader to the existing survey papers~\cite{gui2024survey,zong2023self}. 
%For example, Dong et al. ~\cite{dong2024towards} introduce multimodal pretext tasks, such as Masked Cross-modal Translation and Multimodal Jigsaw Puzzles to improve generalization and open-set detection. Munro and Damen~\cite{munro2020multi} align self-supervised signals across modalities within an adversarial framework, improving fine-grained action recognition.

\subsection{Foundation Models}
Foundation models are large-scale models pre-trained on vast amounts of datasets, enabling effective transfer to various downstream tasks with minimal task-specific supervision. Prominent examples include language models like GPT~\cite{brown2020language}, vision models like SAM~\cite{kirillov2023segment} and DINO~\cite{oquab2023dinov2}, vision-language models like CLIP~\cite{radford2021learning} and Flamingo~\cite{alayrac2022flamingo}, and visual generative models like stable diffusion~\cite{rombach2022high}. For further information on foundation models, we refer the reader to the recent survey paper~\cite{zhou2024comprehensive}. Recent research endeavors aim to enhance DA and DG capabilities using foundation models or adapt them to downstream tasks.

\section{A Generic Framework for Multimodal Adaptation and Generalization}
\label{sec:GenericFramework}

\subsection{Problem Definition} 
We have a labeled source domain $\mathcal{D}_{\mathit{src}}$ and an unlabeled target domain $\mathcal{D}_{\mathit{target}}$, where \(\mathcal{D}_{\mathit{src}} = \{(\mathbf{x}^s_j, y_j)\}_{j=1}^{n_s}\) represents the source domain with \(n_s\) labeled data instances, and \(\mathcal{D}_{\mathit{target}} = \{\mathbf{x}^t_j\}_{j=1}^{n_t}\) denotes the target domain with \(n_t\) data instances.  
Each data instance \(\mathbf{x}_j\) in source and target domain is composed of \(M\) different modalities, expressed as \(\mathbf{x}_j = \{(\mathbf{x}_j)_m \mid m = 1, \dots, M\}\). Labels for both domains are given as \(y_j \in \mathcal{Y} \subset \mathbb{R}\), while labels for the target domain are unavailable during training.
The joint distributions of inputs and labels differ across source and target domains, i.e., \(P^{\mathit{target}}_{XY} \neq P^{\mathit{src}}_{XY}\). 

The goal of \textbf{MMDA} is to learn a robust predictive function \(f: \mathbf{X} \to \mathcal{Y}\) on $\mathcal{D}_{\mathit{src}}$ and $\mathcal{D}_{\mathit{target}}$ that minimizes the prediction error on the unlabeled target domain \(\mathcal{D}_{\mathit{target}}\) under domain shift scenarios:
\begin{equation}
    f = \arg\min_{f} \, \mathbb{E}_{(\mathbf{x},y) \in \mathcal{D}_{\mathit{target}}} [ \ell(f(\mathbf{x}),y) ],
\end{equation}
where \(\mathbb{E}\) denotes the expectation, and \(\ell(\cdot, \cdot)\) is the loss function. 
Given a well-trained multimodal source model $f(\mathbf{x})$ on $\mathcal{D}_{\mathit{src}}$, \textbf{MMTTA} aims to adapt this model online to $\mathcal{D}_{\mathit{target}}$ with streaming data.
The goal of \textbf{MMDG} is to learn a robust and generalizable predictive function $f$ from $\mathcal{D}_{\mathit{src}}$ to achieve a minimum prediction error on $\mathcal{D}_{\mathit{target}}$, where $\mathcal{D}_{\mathit{target}}$ cannot be accessed during training.

\subsection{Generic Framework} 
\begin{figure}[t]
  \centering
  \includegraphics[width=\linewidth]{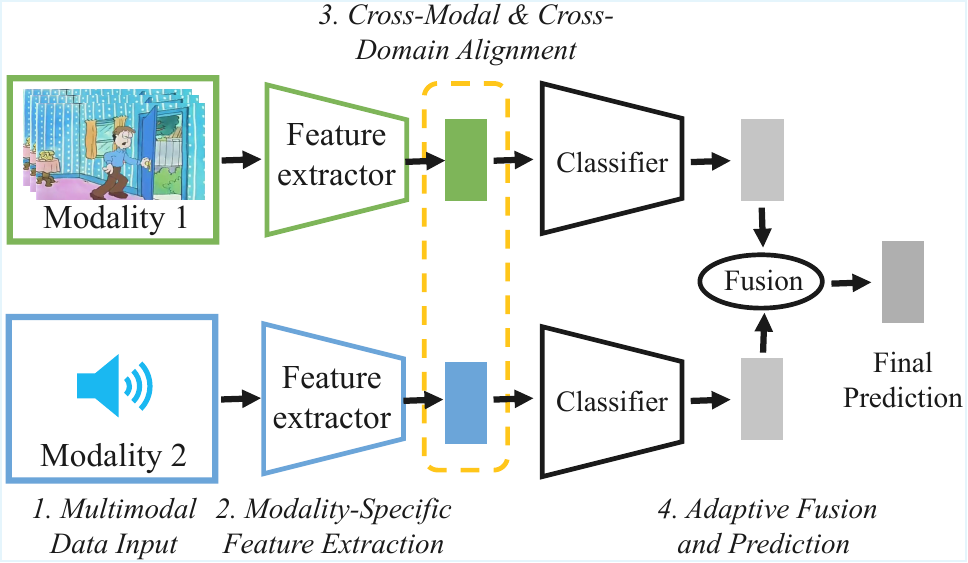}
   \vspace{-0.8cm}
   \caption{{A generic framework for multimodal adaptation and generalization, with four interconnected stages.}}
   \label{fig:gen_frame}
\end{figure}

To provide a structured understanding of the methods discussed in this survey, we propose a generic framework that encapsulates the key components of a typical multimodal DA/DG system. As illustrated in \cref{fig:gen_frame}, this framework is modular and can be adapted to various tasks and scenarios. Our proposed framework consists of four sequential but interconnected stages:

\noindent\textbf{Stage 1: Multimodal Data Input.} The process begins with data from one or more source domains, where each data point consists of multiple modalities  (e.g., video and audio, image and point cloud). For DA, unlabeled data from a target domain is also available.

\noindent\textbf{Stage 2: Modality-Specific Feature Extraction.} Each modality is initially processed by a separate feature extractor to extract high-level feature representations.

\noindent\textbf{Stage 3: Cross-Modal and Cross-Domain Alignment.} This is the heart of any multimodal adaptation or generalization method. The objective is to explicitly reduce the discrepancies that exist both between modalities (the modality gap) and between data distributions across different domains (the domain gap). This stage operates on the features extracted in Stage 2. The techniques employed here can be broadly categorized as feature-level alignment (aligning the feature distributions directly through adversarial learning, discrepancy minimization, contrastive learning, and self-supervised tasks), prediction-level alignment (enforcing consistency between the predictions generated from different modalities through cross-modal interaction), and hybrid alignment (a combination of feature- and prediction-level strategies).

\noindent\textbf{Stage 4: Adaptive Fusion and Prediction.} The final stage aims to combine the aligned multimodal information to produce a robust prediction for the downstream task. The aligned features or predictions from Stage 3 are integrated. Fusion strategies can be categorized as early fusion (concatenating feature vectors before the final classification layer), late fusion (averaging the outputs of separate classifiers for each modality), or hybrid fusion.

\section{Multimodal Domain Adaptation}
\label{sec:mmda}
{Multimodal domain adaptation (MMDA)} aims at adapting a model trained on a labeled source domain to perform effectively on an unlabeled target domain while leveraging multiple modalities of data (e.g., video, audio, and optical flow).  The modality gaps and different levels of distribution shift across modalities make MMDA particularly challenging than unimodal DA. %MMDA utilizes labeled data from the source domain and unlabeled data from the target domain during adaptation. %It combines concepts from domain adaptation and multimodal learning, addressing the challenges of distributional shifts in data across domains while exploiting complementary information from different data modalities.
Existing research on MMDA has primarily focused on two tasks: the action recognition task with video, audio, and optical flow modalities, and the semantic segmentation task with LiDAR point cloud and RGB images. While most proposed methods are partly similar and generally applicable to both tasks, we discuss them separately for clarity.

\subsection{MMDA for Action Recognition} 
In this section, we introduce the most common MMDA methods for action recognition in detail and categorize them into domain-adversarial learning, contrastive learning, and cross-modal interaction.

\subsubsection{Domain-Adversarial Learning}
Adversarial learning-based approaches effectively align multimodal features across domains by leveraging adversarial objectives~\cite{ganin2016domain} to learn domain-invariant representations. Originally designed for unimodal settings, these objectives can be easily extended to multimodal scenarios.  %These methods incorporate domain discriminators into the architecture, where the discriminators are trained to distinguish between domains. Feature extractors are optimized adversarially to confuse the discriminators, promoting robust feature alignment. 
For example, Qi et al.~\cite{U4DA} leverage an adversarial objective to jointly attend and fuse multimodal representation to learn domain-invariant features across modalities.
Differently, MM-SADA~\cite{munro2020multi} incorporates within-modal adversarial alignment alongside multimodal self-supervised alignment for MMDA.%, as shown in~\cref{fig:MM-SADA}. 
Given a binary domain label, $d$, indicating if an example $x \in \mathcal{D}_{\mathit{src}}$ or $x \in \mathcal{D}_{\mathit{target}}$, the domain discriminator for modality $m$ is defined as:
\begin{multline}
    \mathcal{L}^m_{d} = \sum_{x \in \{\mathcal{D}_{\mathit{src}}, \mathcal{D}_{\mathit{target}}\}} -d \log(D^m(F^m(x))) - \\
    (1-d) \log(1-D^m(F^m(x))) ,
    \label{eq:adv}
\end{multline}
where $D^m$ is the domain discriminator for modality $m$ and $F^m$ is the feature extractor. The multimodal self-supervised alignment loss aims to learn the temporal correspondence between modalities and is defined as:
\begin{equation}
    \mathcal{L}_c = \sum_{x \in \{\mathcal{D}_{\mathit{src}}, \mathcal{D}_{\mathit{target}}\}}  -c \log C(F^1(x), ..., F^M(x)) ,
    \label{eq:mmSelf}
\end{equation}
where $C$ is the self-supervised correspondence classifier head and $c$ is a binary label defining if modalities correspond.
Moving beyond simple alignment, Zhang et al.~\cite{9334409} enhance cross-modal collaboration by selecting reliable pseudo-labeled target samples while also addressing missing modality scenarios—where adversarial learning is leveraged to generate absent modalities while preserving semantic integrity.
Yin et al.~\cite{10.1145/3503161.3548313} further extend adversarial learning to temporal sequences, using mix-sample adversarial learning to capture domain-invariant temporal dependencies while dynamically distilling knowledge across modalities to boost adaptability.

% \begin{figure}[t]
%   \centering
%   \includegraphics[width=0.95\linewidth]{imgs/crop_MM-SADA.pdf}
%    \vspace{-0.4cm}
%    \caption{The MM-SADA~\cite{munro2020multi} architecture for MMDA, where adversarial and self-supervised learning are used to align multimodal features.}
%    \label{fig:MM-SADA}
% \end{figure}

\subsubsection{Contrastive Learning}
Contrastive learning~\cite{khosla2020supervised} is a powerful technique for learning transferable representations by pulling positive pairs closer in the feature space while pushing negative pairs apart. In MMDA, it helps align features across both domains and modalities by treating different modalities with the same label as positive pairs.
For instance, Song et al.~\cite{9577559} jointly align clip- and video-level features using self-supervised contrastive learning while minimizing video-level domain discrepancy, thus enhancing category-aware alignment and cross-domain generalization.
Kim et al.\cite{kim2021learning} leverage contrastive learning with modality- and domain-specific sampling strategies by selecting multiple positive and negative samples to jointly regularize cross-modal and cross-domain feature representations.

\subsubsection{Cross-Modal Interaction}
Cross-modal interaction methods enhance multimodal feature learning by fostering information exchange between modalities during adaptation, enabling models to capture complementary and interdependent relationships across modalities.
For instance, Lv et al.~\cite{10.1145/3474085.3475660} model modality-specific classifiers as teacher-student sub-models, using prototype-based reliability measurement for adaptive teaching and asynchronous curriculum learning, and employing reliability-aware fusion for robust final decisions.
Huang et al.~\cite{10.1145/3503161.3548009} address source-free MMDA by leveraging self-entropy-guided Mixup~\cite{zhang2018mixup} to generate synthetic samples and aligning these with hypothetical source-like samples using multimodal and temporal relative alignment.
Zhang et al.~\cite{zhang2022audio} assume a different perspective by proposing an audio-adaptive encoder and an audio-infused recognizer to tackle domain shifts in action recognition across scenes, viewpoints, and actors. By leveraging domain-invariant audio activity information, they refine visual representations through absent activity learning and enhance recognition with visual cues.
Yang et al.~\cite{9879925} demonstrate that enhancing the transferability of each modality through cross-modal interaction prior to performing cross-domain alignment is more effective than directly aligning the multimodal inputs.
Recently, Dong et al.~\cite{dong2024towards} addressed  MMDA in an open-set setting  by designing two self-supervised tasks -- masked cross-modal translation and multimodal Jigsaw puzzles -- to learn robust multimodal features for improved  generalization and open-class detection. Additionally, they incorporated  an entropy weighting mechanism to effectively balance modality-specific losses.

%\input{tables/MMDASS}
%%%%%%%%%%%%%%%%%%%%%%%%%%%%%%%%%%%%%%%%%%
\subsection{MMDA for Semantic Segmentation}
In this section, we introduce most common MMDA methods for semantic segmentation in detail and categorize them into xMUDA and its extensions, domain-adversarial learning, and cross-modal interaction.

\subsubsection{xMUDA and Its Extensions}
\begin{figure}[t]
  \centering
  \includegraphics[width=\linewidth]{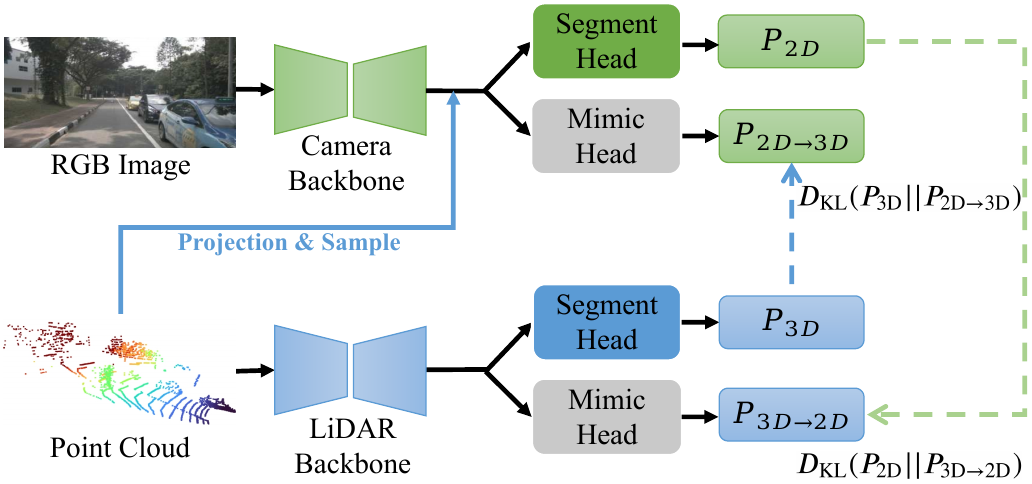}
   \vspace{-0.8cm}
   \caption{The xMUDA~\cite{jaritz2020xmuda} architecture for MMDA, which promotes cross-modal prediction consistency through multi-head mutual mimicking.}
   \label{fig:xMUDA}
\end{figure}

Jaritz et al.~\cite{jaritz2020xmuda} introduce the first MMDA framework named xMUDA for 3D semantic segmentation (3DSS), promoting cross-modal prediction consistency through multi-head mutual mimicking (\cref{fig:xMUDA}). An unsupervised cross-modal divergence loss is applied to both the source and target domains to ensure effective cross-modal alignment:
\begin{align}
\label{eq:kl_xmuda}
\mathcal{L}_{xM} &= D_{KL}(P_x^{(n,c)} \| Q_x^{(n,c)}) \notag \\
          &= -\frac{1}{N} \sum_{n=1}^{N} \sum_{c=1}^{C} P_x^{(n,c)} \log \frac{P_x^{(n,c)}}{Q_x^{(n,c)}},
\end{align}
where \((P, Q) \in \{(P_{2D}, P_{3D \rightarrow 2D}), (P_{3D}, P_{2D \rightarrow 3D})\}\), $N$ the number of 3D points and $C$ is the number of classes. Here, \(P\) denotes the target distribution from the main prediction, while \(Q\) represents the mimicking prediction used to approximate \(P\). xMUDA also has a variant xMUDA$_{\text{PL}}$, which leverages  pseudo-labels for self-training and serves as a strong baseline for MMDA. As a pioneering work, xMUDA introduced  a new benchmark using nuScenes~\cite{caesar2020nuscenes}, A2D2~\cite{geyer2020a2d2}, and SemanticKITTI~\cite{behley2019semantickitti} datasets, covering  three adaptation scenarios: day-to-night,  country-to-country, and dataset-to-dataset. Many subsequent studies have built upon  xMUDA, extending it from different perspectives. % using advanced data augmentation, fusion methods, or cross-modal interaction.
%xMoSSDA~\cite{jaritz2022cross} extends xMUDA to semi-supervised settings, demonstrating superior performance.

\noindent\textbf{Extension via Data Augmentation.}
Data augmentation techniques have been explored to enhance cross-modal alignment in xMUDA. For example, Li et al.~\cite{10.1145/3503161.3547990} propose a multimodal style transfer strategy and a target-aware teacher framework to perform cross-domain and cross-modal knowledge distillation on source and synthesized target-style data. 
Complementing this, Chen et al.~\cite{10.1145/3581783.3612320} employ CutMix~\cite{9008296} and Mix3D~\cite{nekrasov2021mix3d} to augment 2D and 3D training data, facilitating 2D-3D interaction and intra-domain cross-modal learning. %CoMoDaL simultaneously incorporates intra-domain cross-modal guidance and inter-modal cross-domain alignment while eliminating the dependency on source 3D data in the xMUDA setting.
Recently, Cao et al.~\cite{cao2024mopa} enhanced  xMUDA's pipeline by incorporating 3D rare objects collected from real-world scenarios and leveraging pixel-wise supervision from the SAM~\cite{kirillov2023segment} model. This approach effectively addresses imbalanced supervision and significantly improves the segmentation of rare objects.

\noindent\textbf{Extension via Fusion.}
Beyond augmentation, fusion-based strategies refine xMUDA by improving information exchange between modalities. For instance, Wu et al.~\cite{10.1145/3581783.3612013} perform cross-modal and cross-domain alignments through knowledge distillation using fused cross-modal representations, maximizing correlation and complementarity between heterogeneous modalities to mitigate domain shift.
Cardace et al.~\cite{cardace2023exploiting} further strengthen fusion by feeding depth features into the 2D branch while dynamically enriching the 3D network with RGB features. By employing middle fusion across both branches, they effectively exploit intrinsic cross-modal complementarity.
Taking a different approach, Simons et al.~\cite{10377604}  introduce a dynamic selection mechanism for fused and unfused rectified pseudo-labels, enabling  self-training in source-free MMDA for 3DSS.

\noindent\textbf{Extension via Cross-modal Interaction.}
Zhang et al.~\cite{10.1145/3503161.3547987} introduce  plane-to-spatial and discrete-to-textured self-supervised tasks to train  models in a mixed-domain setting,enhancing modality-specific learning and mitigating domain shift. %A domain-category adversarial learning is proposed to align domains in a category-wise manner to maintain the discriminability of category features.
Xing et al.~\cite{xing2023cross} strengthen  xMUDA by incorporating cross-modal contrastive learning and a neighborhood feature aggregation module, reinforcing  2D-3D consistency across domains while capturing richer contextual information.
Building on this, Zhang et al.~\cite{zhang2023mx2m} integrate  masked cross-modal modeling to bridge  large domain gaps and introduce dynamic cross-modal filters for feature matching. This enables the model  to dynamically leverage  2D-3D complementarity, improving overall robustness and adaptability. 

\subsubsection{Domain-Adversarial Learning}
Similar to action recognition, domain-adversarial learning methods for 3DSS leverage adversarial objectives to learn domain-invariant representations. For instance, Peng et al.~\cite{peng2021sparse} introduce sparse-to-dense feature alignment, enforcing  intra-domain point-pixel correspondence while employing  adversarial learning across both domains and modalities for inter-domain alignment. % making it the first approach to address cross-modal learning at both levels. 
In contrast, Liu et al.~\cite{LIU2021211} focus adversarial learning on the image modality and propose a threshold-moving strategy to mitigate data imbalance during inference.
Beyond  pure adversarial alignment, Man et al.~\cite{man2023dualcross}  introduce a distillation framework that transfers knowledge from a LiDAR teacher model to a camera student model through  feature supervision on depth estimation and Bird's-Eye View (BEV) embeddings. Additionally, multi-stage adversarial learning further refines  feature alignment  across domains.% enabling accurate monocular 3D perception under significant domain shifts.
%DualCross~\cite{man2023dualcross} addresses mixed domain and modality mismatches in 3D domain adaptation for BEV perception, where the source domain provides annotated LiDAR-camera pairs, and the target domain contains only unannotated camera images. It introduces a distillation framework that transfers knowledge from a LiDAR teacher model to a camera student model via feature supervision on depth estimation and BEV embeddings. Multi-stage adversarial learning further aligns feature spaces across domains, enabling accurate monocular 3D perception under significant domain shifts.

\subsubsection{Cross-Modal Interaction}
To enhance  cross-modal interaction, Vobecky et al.~\cite{vobecky2022drive} propose  a cross-modal unsupervised approach for 2D semantic segmentation (2DSS) using unannotated paired LiDAR and camera data. Their method  extracts 3D-consistent object segments based on geometrical properties, then applies projection and clustering to generate 2D pseudo-ground truth, enabling knowledge distillation with cross-modal spatial constraints. %This highlights the potential of leveraging unlabeled paired data for MMDA.   
Yin et al.~\cite{10376488} address source-free MMDA for 2DSS by integrating a multimodal auxiliary network during training. They employ middle fusion and enforce prediction consistency between augmented depth-RGB pairs to facilitate  cross-modal learning.  %Additionally, it leverages the prototypical pseudo-label denoising method~\cite{zhang2021prototypical} for further enhanced adaptation.
Rizzoli et al.~\cite{rizzoli2024source} further enhance multimodal learning by integrating depth data into a vision transformer at multiple levels --input, feature, and output stages. Their approach employs color and depth style transfer for early domain alignment while cross-modal self-attention generates enriched  feature representations  for improved  semantic extraction.  %A depth-based entropy minimization strategy is proposed that adaptively weights regions by distance, further improving adaptation.
% Bultmann et al.~\cite{BULTMANN2023104286} enable real-time semantic inference and fusion of LiDAR, RGB, and thermal sensor modalities for both semantic segmentation and object detection. Their method employs late fusion and label propagation,  allowing  adaptation across sensors and domains for robust multimodal domain adaptation.
%address domain adaptation between autonomous driving and UAVs (with an increased vertical field of view) for point cloud segmentation by retraining the LiDAR backbone using pseudo-labels derived from aggregated semantic maps (RGB and thermal modalities) and target domain datasets. The refined model employs late fusion and spatio-temporal aggregation to integrate image and point cloud segmentation, achieving fully and accurately annotated semantic point clouds, validated across diverse urban environments and challenging disaster test sites.

% BEV perception, LiDARTeacher and Camera-Student knowledge distillation model, adversarial

\subsection{MMDA for Other Tasks}
Beyond action recognition and semantic segmentation, MMDA has been explored across a range of diverse tasks.%, such as sentiment analysis, medical image segmentation, etc.

\noindent\textbf{Sentiment Analysis and Emotion Recognition.}
% For example, Ma et al.~\cite{ma2019deep} address MMDA for cross-domain object and event recognition by leveraging  stacked attention to capture  semantic representations while  applying multi-channel constraints to enhance category discrimination. 
Sentiment analysis and emotion recognition in modern digital media are inherently multimodal, often involving text (e.g., a tweet, a product review) and an accompanying image or video.
Liu et al.~\cite{10204066} address MMDA for sentiment analysis by introducing a tensor-based alignment module to model  relationships between domains and modalities, along with a dynamic domain generator that creates transitional samples. %Their approach achieves state-of-the-art  performance in multimodal sentiment analysis and video text classification tasks.
Zhang et al.~\cite{zhang-etal-2024-amanda} address MMDA for emotion recognition by independently learning optimal representations for each modality while adaptively balancing domain alignment across modalities through dynamic weighting, ensuring more effective cross-modal adaptation.

\noindent\textbf{Cross-Modal Retrieval.} The goal of cross-modal retrieval is to retrieve relevant items from one modality (e.g., images) given a query from another (e.g., text). Liu et al.~\cite{liu2021adaptive} force the model to preserve common semantic information across domains while discarding domain-specific noise, enabling effective retrieval even without target-domain pairs.

\noindent\textbf{Medical Image Analysis.}
In modern clinical practice, diagnosis and treatment planning often rely on integrating information from multiple imaging modalities. Yao et al~\cite{9741336} address cross-modality medical image segmentation by introducing multi-style image translation for better domain alignment.

\noindent\textbf{Generative Modeling.} Generative domain adaptation aims to adapt a pre-trained generative model, such as diffusion models, to synthesize data from a new target domain. Zhao et al.~\cite{zhao2025doracycle} integrate two parallel cycles: a text-to-image-to-text cycle and an image-to-text-to-image cycle. The model is optimized by enforcing that the output of a cycle matches its input.

\subsection{Summary and Insights for MMDA}
MMDA has emerged as a powerful framework for leveraging multiple modalities to bridge the gap between source and target domains. While significant progress has been made in tasks like action recognition and semantic segmentation, MMDA faces unique challenges, such as modality imbalance, missing modalities, and the need for effective cross-modal alignment. Current approaches, including domain-adversarial learning, contrastive learning, and cross-modal interaction, have demonstrated success in addressing these challenges by learning domain-invariant representations and fostering modality-specific and modality-shared knowledge transfer. %However, the field still grapples with scalability to high-dimensional modalities (e.g., 3D point clouds or video streams) and the computational cost of aligning multiple modalities in real-world applications.
Future research could explore modality-agnostic adaptation frameworks that dynamically prioritize the most informative modalities during adaptation, reducing redundancy and computational overhead. %Additionally, self-supervised learning techniques tailored for MMDA could further enhance generalization by leveraging unlabeled target domain data more effectively. 
Another promising direction is the integration of causal inference to model the underlying relationships between modalities and domains, enabling more robust adaptation under distribution shifts. Finally, as multimodal datasets grow in size and complexity, scalable and efficient MMDA algorithms that can handle large-scale, heterogeneous data will be critical. %By addressing these challenges, MMDA has the potential to enable more robust and adaptive multimodal systems in real-world applications, such as autonomous driving, healthcare, and human-computer interaction.

\section{Multimodal Test-time Adaptation}
\label{sec:mmtta}
In contrast to multimodal domain adaptation where both source and target domain data are available during adaptation, {multimodal test-time adaptation (MMTTA)} aims to adapt a pre-trained source model online to a target domain without having access data from the source domain.

% \subsection{Problem Definition}
% Let \(\mathcal{D}_{\mathit{src}} = \{(\mathbf{x}^s_j, y_j)\}_{j=1}^{n_s}\) represent  the source domain dataset which follows the distribution $P^{\mathit{src}}_{XY}$ and each sample $\mathbf{x}_j$  consists of $M$ modalities, denoted as $\mathbf{x}_j=\{x_j^m \mid m=1,\cdots,M\}$. Similarly, let \(\mathcal{D}_{\mathit{target}} = \{\mathbf{x}^t_j\}_{j=1}^{n_t}\) represent the target domain dataset with distribution $P^{\mathit{target}}_{XY}$. The label spaces for both domains are given as \(y_j \in \mathcal{Y} \subset \mathbb{R}\). Let $f: \mathbf{X} \to \mathcal{Y}$ denote a neural network trained on the source distribution $P^{\mathit{src}}_{XY}$. In MMTTA, $f$ consists of $M$ feature extractors $g_m(\cdot)$ and a classifier $h(\cdot)$. Each feature extractor $g_m(\cdot)$  processes modality $m$ to produce an embedding~$\mathbf{E}^{m}$, and the classifier $h(\cdot)$ combines these embeddings to generate a prediction probability $\hat{p}$:
% \begin{align}
%   \hat{p} = \delta(f(\mathbf{x}))  \notag 
%           &= \delta(h([g_1(x^1), ..., g_M(x^M)]))  \notag \\
%           &= \delta(h([\mathbf{E}^{1}, ..., \mathbf{E}^{M}])) .
%   \label{eqn:pred2}
% \end{align}
% \noindent 
% where $\delta(\cdot)$ denotes the softmax function. 
% Given a well-trained multimodal source model $f(\mathbf{x})$ on $\mathcal{D}_{\mathit{src}}$, MMTTA aims to adapt this model online to the unlabeled target domain $\mathcal{D}_{\mathit{target}}$, where $P^{\mathit{target}}_{XY} \neq P^{\mathit{src}}_{XY}$.

\subsection{Methods for Multimodal Test-time Adaptation}
MMTTA is a relatively new research direction, with only a limited number of studies addressing it. Existing research primarily explores MMTTA  in action recognition, semantic segmentation, and other tasks.

\subsubsection{MMTTA for Action Recognition}
READ, proposed by Yang et al.~\cite{yang2023test}, addresses MMTTA under reliability bias, where modality-specific information discrepancies arise  from intra-modal distribution shifts. Unlike previous TTA methods previous TTA methods~\cite{wang2021tent,niu2022efficient}, which update batch normalization statistics and transformation parameters, READ takes a different approach by dynamically modulating cross-modal attention in a self-adaptive manner  to ensure reliable fusion. Additionally, READ  introduces a novel confidence-aware loss function $\mathcal{L}_{ra}$, designed to enhance the robustness of multimodal adaptation:
\begin{equation}
  \mathcal{L}_{ra}
  =\frac{1}{B} \sum_{i=1}^{B}  p_i \log{\frac{e\gamma}{p_i}} ,
  \label{eqn:ra}
\end{equation}
where $B$ is the batch size, $p_i$ is the confidence of the prediction $\hat{p}_i$, i.e., $p_i=\max(\hat{p}_i)$, and $\gamma$ is a threshold for confident prediction. $\mathcal{L}_{ra}$ helps the model focus more on the high-confident prediction while preventing the noise from the low-confident predictions.
In a different line of research, Xiong et al.~\cite{xiong2024modality} propose a teacher-student memory bank framework  combined with self-assembled source-friendly feature reconstruction to align multimodal prototypes effectively. Their approach mitigates domain shifts in MMTTA by preserving cross-modal consistency and enhancing feature adaptability.
Furthermore, Lei et al.~\cite{lei2024twolevel} adopt a two-level objective function  that incorporates Shannon entropy loss and a diversity-promoting loss. This approach effectively addresses both intra-modal distribution shifts and cross-modal reliability bias within the modality fusion block, ensuring more robust multimodal adaptation.
Beyond domain shifts, Dong et al.~\cite{dong2025aeo} extend MMTTA to the open-set setting, where previously unseen categories emerge  during test-time adaptation. They propose adaptive entropy-aware optimization (AEO), a novel approach that  amplifies the entropy difference between known and unknown samples during online adaptation. AEO consists of two key components: (1) unknown-aware adaptive entropy optimization and (2) adaptive modality prediction discrepancy optimization. The unknown-aware adaptive entropy optimization module adaptively weights and optimizes each sample based on its prediction uncertainty and is defined as:
\begin{equation}
W_{ada} =  \tanh(\beta\cdot(H(\hat{p})-\alpha)),
\label{eq_w}
\end{equation}
\begin{equation}
\mathcal{L}_{AdaEnt} =  -H(\hat{p}) W_{ada},
\label{eq_adaent}
\end{equation}
where $\tanh$ is the hyperbolic tangent function, $W_{ada}$ is the adaptive weight assigned to each sample, $H(\hat{p})$ is the normalized entropy of prediction $\hat{p}$, computed as $H(\hat{p}) = -(\sum_c \hat{p}_c \log \hat{p}_c)/\log(C)$,  with 
$C$ being the number of classes. 

The adaptive modality prediction discrepancy optimization module optimizes the prediction discrepancy  across modalities and is defined as:
\begin{equation}
\mathcal{L}_{AdaDis} =  - (Dis(\hat{p}^1, \hat{p}^2)) W_{ada},
\label{eq_adadis}
\end{equation}
where $W_{ada}$ is the adaptive weight calculated in~\cref{eq_w}. 
In addition to the pure open-set TTA scenarios, AEO has also demonstrated effectiveness and versatility in challenging long-term and continual adaptation scenarios.

\subsubsection{MMTTA for Semantic Segmentation}
Beyond  action recognition, MMTTA has also been applied to 3D semantic segmentation and other tasks. For instance, Shin et al.~\cite{shin2022mm} propose an intra-modal pseudo-label Generation module, which independently  generates pseudo-labels for each modality, and an inter-modal pseudo-label refinement module, which adaptively selects and refines pseudo-labels across modalities to enhance cross-modal consistency and improve adaptation. %The intra-PG module combines predictions from slowly updated and fast-updated models to produce robust pseudo-labels, while the inter-PR module evaluates prediction consistency between these models to confidently select pseudo-labels.
Building on this, Cao et al.~\cite{cao2023multi} explore multi-modal continual test-time adaptation, addressing dynamically evolving domains over time. It facilitates dynamic adaptation for 3D semantic segmentation by attending to reliable modalities and mitigating catastrophic forgetting through dual-stage mechanisms and class-wise momentum queues designed for continual domain shifts.
Recently, Cao et al.~\cite{cao2024reliable} further enhanced 3D segmentation by leveraging reliable spatial-temporal correspondences, filtering unreliable predictions, and employing cross-modal learning to maintain consistency across consecutive frames.

\subsubsection{MMTTA for Other Tasks}
Park et al.~\cite{park2024test} focus on MMTTA for depth completion using a single image and an associated sparse depth map. They reduce the domain gap by employing a source-trained embedding module that aligns image and sparse depth features from the target domain in a single pass. % achieving improved performance without requiring access to source data.
Wang et al.~\cite{wang2024heterogeneous} tackle MMTTA for person re-identification, enhancing model generalization by leveraging relationships among heterogeneous modalities. %They incorporate a cross-identity inter-modal margin loss and employs a multimodal test-time training strategy for self-supervised adaptation.
Recently, Li et al~\cite{li2024test} address MMTTA for cross-modal retrieval that refines query predictions and employs a joint objective to mitigate the effects of query shift.
Besides, MMTTA has also been applied to diverse domains such as sleep stage classification~\cite{jia2024atta}, sentiment analysis~\cite{guo2025bridging}, and color-thermal semantic segmentation~\cite{liu2023test}

% \subsection{Evaluation of MMTTA}

% Evaluation of MMTTA algorithms often follows the settings of unimodal TTA. Given a multimodal dataset containing at least two distinct domains, one or multiple of them are used as source domains for model training, while the rest are treated as target domains. At test time, the model adapts to the target domain in an online manner using the test data itself, without accessing the source domain or target labels. The average performance on the target domains is always concerned and evaluated. The model is usually selected based on training-domain validation, which holds out a subset of the source domain data for model selection.

\subsection{Summary and Insights for MMTTA}
The emerging field of MMTTA has shown promising results across various tasks, including action recognition, semantic segmentation, and depth completion. Current methods primarily focus on addressing challenges such as intra-modal distribution shifts, cross-modal reliability bias, and open-set adaptation. Techniques like dynamic cross-modal attention modulation, pseudo-label refinement, and entropy-aware optimization have demonstrated effectiveness in improving robustness and adaptability. However, several open challenges remain, such as scaling MMTTA to more complex multimodal tasks, handling extreme domain shifts, and ensuring efficient real-time adaptation. Future research could explore the integration of MMTTA with foundation models, leveraging large-scale pretraining to enhance generalization, and developing unified frameworks that can seamlessly adapt to diverse multimodal scenarios. Additionally, investigating the theoretical underpinnings of MMTTA, such as the interplay between modality-specific and cross-modal learning, could provide deeper insights into its mechanisms and limitations. %As MMTTA continues to evolve, it holds significant potential to enable more reliable and adaptive multimodal systems in real-world applications.

\section{Multimodal Domain Generalization}  
\label{sec:mmdg}  
In contrast to multimodal domain adaptation and test-time adaptation, {multimodal domain generalization (MMDG)} presents a more challenging problem setting. In MMDG, the model is trained only on source domains with multiple modalities to generalize across unseen domains, without prior exposure to target domain data during training, making it harder than MMDA and MMTTA.

\subsection{Methods for Multimodal Domain Generalization} %Taxonomy on MMDG Algorithms %Methodology
Similar to MMTTA, MMDG is also a relatively new research direction, with only a few studies addressing this challenging problem in action recognition and semantic segmentation tasks.

\subsubsection{MMDG for Action Recognition}
Planamente et al.~\cite{planamente2022domain} propose  the first MMDG approach for egocentric activity recognition, introducing the relative norm alignment loss, which aligns the mean feature norms across different modalities:
\begin{equation}\label{formula:rna}
\mathcal{L}_{align}=\left(\frac{\mathbb{E}[\mathbf{E}^{v}]}{\mathbb{E}[\mathbf{E}^{a}]} - 1\right)^2,
\end{equation}
where $\mathbb{E}[\mathbf{E}^{m}]$ is the mean feature for the $m$-th modality in each batch. This loss  prevents the dominance of a single modality during multimodal joint training, enhancing generalization  generalization across domains. Building on this, a subsequent study by Planamente et al.~\cite{planamente2024relative} refines  the relative norm alignment loss by extending it to align class-level feature norms, further improving its effectiveness  in multimodal domain generalization. %Both approaches can also effectively adapt to the MMDA task by applying unsupervised $\mathcal{L}_{RNA}$ to the unlabeled target domain data.

\begin{figure}[t]
  \centering
  \includegraphics[width=\linewidth]{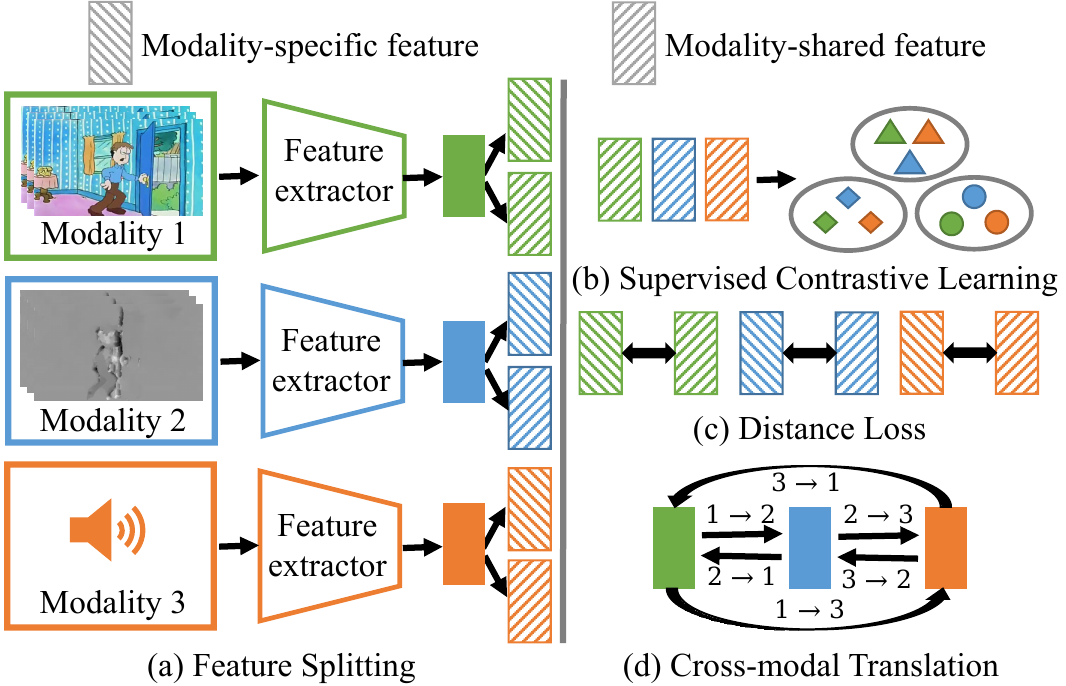}
   \vspace{-0.8cm}
   \caption{The SimMMDG~\cite{dong2023simmmdg} architecture for MMDG, which separates features into modality-specific and modality-shared components for better generalization.}
   \label{fig:SimMMDG}
\end{figure}

Recently, Dong et al.~\cite{dong2023simmmdg} proposed SimMMDG, a unified framework to address MMDG across various  scenarios, including multi-source, single-source, and missing-modality settings, as illustrated  in~\cref{fig:SimMMDG}. SimMMDG introduces a feature disentanglement strategy that decomposes  features into modality-specific and modality-shared components, enhancing generalization across domains. Specifically, given a unimodal embedding $\mathbf{E}$, SimMMDG splits it into $\mathbf{E} = [\mathbf{E}_s; \mathbf{E}_c]$, where $\mathbf{E}_s$ is a modality-specific feature and $\mathbf{E}_c$ is a modality-shared feature. This  disentanglement is enforced through supervised contrastive learning~\cite{NEURIPS2020_d89a66c7} on modality-shared features, combined with distance constraints on modality-specific features to encourage  diversity, ensuring robust multimodal domain generalization. For a set of $N$ randomly sampled label pairs in a batch, $\{\boldsymbol{x}_j,y_j\}_{j=1, ..., N}$, the corresponding batch used for training consists of $M \times N$ pairs, $\{\boldsymbol{\tilde{x}}_k,\tilde{y}_k\}_{k=1, ..., M \times N}$, where $\boldsymbol{\tilde{x}}_{M \times j}$, $\boldsymbol{\tilde{x}}_{M \times j-1}$, ... , $\boldsymbol{\tilde{x}}_{M \times j-M+1}$ are data instances from $M$ different modalities in $\boldsymbol{x}_j$ ($j=1, ..., N$) and $\tilde{y}_{M \times j}=\tilde{y}_{M \times j-1}=...=\tilde{y}_{M \times j-M+1}=y_j$. Let $i\in I\equiv\{1, ..., M \times N\}$ be the index of an arbitrary unimodal sample within a batch. We define $A(i)\equiv I\setminus\{i\}$, $P(i)\equiv\{p\in A(i):{\tilde{y}}_p={\tilde{y}}_i\}$ as the set of indices of all positive samples in the batch which share the same label as $i$. The cardinality of $P(i)$ is denoted as $|P(i)|$. The multimodal supervised contrastive learning loss can be written as: 
\begin{equation}
  \mathcal{L}_{con}
  =\sum_{i\in I}\frac{-1}{|P(i)|}\sum_{p\in P(i)}\log{\frac{\text{exp}\left(\boldsymbol{z}_i\bigcdot\boldsymbol{z}_p/\tau\right)}{\sum\limits_{a\in A(i)}\text{exp}\left(\boldsymbol{z}_i\bigcdot\boldsymbol{z}_a/\tau\right)}} ,
  \label{eqn:supervised_loss}
\end{equation}
with $\boldsymbol{z}_k=Proj(g(\boldsymbol{\tilde{x}}_k))\in\mathcal{R}^{D_P}$, where $g(\cdot)$ is the feature extractor, that maps $\boldsymbol{x}$ to modality-specific and modality-shared features, $\mathbf{E} = [\mathbf{E}_s; \mathbf{E}_c] = g(\boldsymbol{x})$, where $\mathbf{E}_s, \mathbf{E}_c \in R^{D_E}$, 
%$\mathbf{E}_c=g(\boldsymbol{x})\in\mathcal{R}^{D_E}$
and $Proj(\cdot)$ is the projection network that maps $\mathbf{E}_c$ to a vector $\boldsymbol{z}=Proj(\boldsymbol{\mathbf{E}_c})\in\mathcal{R}^{D_P}$.
The inner product between two projected feature vectors is denoted by $\bigcdot$, and $\tau\in\mathcal{R}^+$ is a scalar temperature parameter. The distance loss is proposed to ensure that the modality-specific features $\mathbf{E}_s$ carry unique and complementary information and is defined as:
\begin{equation}
  \mathcal{L}_{dis}
  =\frac{-1}{M} \sum_{i=1}^{M} ||\mathbf{E}_s^{i} - \mathbf{E}_c^{i}||_2^2,
  \label{eqn:dis_loss}
\end{equation}
where $M$ is the number of modalities, $\mathbf{E}_s^{i}$ and $\mathbf{E}_c^{i}$ are the modality-specific and modality-shared features of the $i^\mathrm{th}$ modality.
Finally, a cross-modal translation module is further proposed to ensure the meaningfulness of modality-specific features and improve robustness in missing-modality scenarios:
\begin{equation}
  \mathcal{L}_{trans}
  =\frac{1}{M(M-1)} \sum_{i=1}^{M} \sum_{j \neq i} ||MLP_{\mathbf{E}^{i} \rightarrow \mathbf{E}^{j}}(\mathbf{E}^{i}) - \mathbf{E}^{j}||_2^2 ,
  \label{eqn:trans_loss}
\end{equation}
where MLP is a multi-layer perception to translate the embedding of the $i^\mathrm{th}$ modality to the $j^\mathrm{th}$ modality.

Building on SimMMDG, MOOSA~\cite{dong2024towards} extends MMDG to the open-set setting for the first time, introducing the multimodal open-set domain generalization problem. MOOSA leverages self-supervised tasks, including  masked cross-modal translation and multimodal Jigsaw puzzles, to enhance generalization, while  an entropy-weighting mechanism  balances losses across modalities for improved adaptation.
In contrast, Fan et al.~\cite{fan2024crossmodal} identify modality competition and discrepant uni-modal flatness as key challenges  in MMDG. To address these issues, they propose constructing consistent flat loss regions and improving  cross-modal knowledge transfer, ensuring better knowledge exploitation for each modality and mitigating the imbalance in multimodal learning.

\subsubsection{MMDG for Semantic Segmentation}
Beyond action recognition, MMDG has been applied to 3D semantic segmentation, as demonstrated in BEV-DG~\cite{li2023bev}. BEV-DG employs a BEV-driven domain contrastive learning strategy to optimize the extraction  of domain-irrelevant representations.  Additionally, it  introduces a BEV-based area-to-area fusion mechanism to improve cross-modal learning, enabling more robust feature alignment. %These innovations offer greater robustness and fault tolerance compared to traditional point-level alignment methods.

\subsubsection{MMDG for Other Tasks}
Beyond action recognition and semantic segmentation, MMDG has also been applied to diverse domains such as hyperspectral image classification~\cite{10095723}, wetland classification~\cite{10806800}, medical image segmentation~\cite{jiang2024dfsegmentation}, and face anti-spoofing~\cite{ma2025denoising}.

%Learning Scenarios of MMDG algorithms
%APPLICATIONS %3D Semantic Segmentation %Action Recognition
%DATASETS, EVALUATION, AND BENCHMARK

% \subsection{Evaluation of MMDG}
% Evaluation of MMDG algorithms often follows the settings of unimodal DG~\cite{zhou2022dg}. Given a multimodal dataset containing at least two distinct domains, one or multiple of them are used as source domains for model training while the rest are treated as target domains. A model learned from the source domains is directly tested in the target domains without any form of adaptation and the average performance in target domains is always concerned and evaluated. The model is usually selected based on training-domain validation, which holds out a subset of training data for model selection.

\subsection{Summary and Insights for MMDG}
MMDG represents a significant advancement in enabling multimodal models to generalize across unseen domains without access to target domain data during training. Current approaches, such as feature norm alignment, feature disentanglement, and self-supervised learning, have demonstrated promising results in tasks like action recognition and semantic segmentation. However, MMDG remains a challenging problem due to issues like modality competition, discrepant unimodal flatness, and the need for robust cross-modal alignment. Future research could explore the integration of foundation models and large-scale pretraining to enhance generalization capabilities further. Additionally, investigating theoretical frameworks to better understand the interplay between modality-specific and modality-shared features could provide deeper insights into MMDG mechanisms. %As the field progresses, addressing challenges such as scalability to more complex tasks, handling extreme domain shifts, and ensuring efficient training will be critical. MMDG holds immense potential for real-world applications, particularly in scenarios where target domain data is unavailable or constantly evolving, paving the way for more adaptive and robust multimodal systems.

\section{Domain Adaptation and Generalization with the Help of Multimodal Foundation Models}
\label{sec:dadgf}

With the recent emergence of large-scale pre-trained multimodal foundation models (MFMs) such as CLIP~\cite{radford2021learning}, stable diffusion~\cite{rombach2022high}, and segment anything model (SAM)~\cite{kirillov2023segment}, numerous studies have explored leveraging these models to enhance generalization capabilities. These approaches can be categorized into three main directions: data augmentation, knowledge distillation, and learning strategies. %\cref{fig-dafm} provides an overview of representative methods.

% \begin{figure}[t!]
% 	\centering
% 	\resizebox{.5\textwidth}{!}{
% 	\input{imgs/fig-DAFM}
% 	}
%    \vspace{-0.7cm}
% 	\caption{Taxonomy of Domain Adaptation and Generalization with MFMs.}
% 	\label{fig-dafm}
% \end{figure}

\subsection{Multimodal Foundation Models}
MFMs are large-scale machine learning models designed to process and integrate multiple types of modalities, such as text, images, audio, and video, to generate meaningful representations and perform diverse tasks. These models are typically pre-trained on vast datasets using self-supervised or weakly supervised learning techniques and can be adapted to various downstream applications through fine-tuning, prompting, or other strategies.

\noindent\textbf{Contrastive Language–Image Pre-Training (CLIP)~\cite{radford2021learning}} is a vision-language model comprising an image encoder that maps high-dimensional images to a low-dimensional embedding space and a text encoder that generates text representations from natural language. CLIP, trained on $400$ million image-text pairs, aligns image and text embedding spaces using contrastive loss. For a batch of image-text pairs, CLIP maximizes the cosine similarity for matched pairs while minimizing it for unmatched pairs. During testing, the class names of a target dataset are embedded using the text encoder with prompts in the form of “a photo of a [CLASS]”, where the class token is replaced with specific class names, such as “cat”, “dog” or “car”. The text encoder generates text embeddings $\mathbf{T}_c$ for each class, and the prediction probability for an input image $\mathbf{x}$, with embedding~$\mathbf{I}_{\mathbf{x}}$, is computed as:
\begin{equation} \label{eqn:clip-metric}
    P(y|\mathbf{x}) = \frac{\exp{\big( \cos \left(\mathbf{I}_\mathbf{x}, \mathbf{T}_{y} \right) / \tau \big)} } { \sum_{c=1}^{C} \exp{\big( \cos \left(\mathbf{I}_\mathbf{x}, \mathbf{T}_{c} \right) / \tau \big)} },
\end{equation}
where $\cos(\cdot,\cdot)$ is the cosine similarity between embeddings, and $\tau$ is a temperature. Recent works on DA and DG use CLIP's text encoder to guide the generation of diverse visual features or distill CLIP's visual encoder into a smaller student model for better generalization.

\noindent\textbf{Diffusion Models}~\cite{rombach2022high}, such as denoising diffusion probabilistic models \cite{ho2020denoising}, learn the desired data distribution through a Markov chain of length $T$.
In the forward pass, noise is progressively added to a data sample $x_0$ to create a sequence of noisy samples $x_t, t\in T$.
In the reverse process, a model $\epsilon_\theta$, parameterized by $\theta$, predicts the added noise at each step $t$.
Stable diffusion~\cite{rombach2022high} applies this denoising process to the latent representation $z$ of $x$ in the latent space of VQGAN~\cite{esser2020taming} with the learning objective of predicting the added noise at each time step $t$ as:
\begin{equation}
\mathcal{L}=\mathbb{E}_{\mathcal{E}(x), \epsilon \sim \mathcal{N}(0,1), t}\left[\left\|\epsilon-\epsilon_\theta\left(z_t, t\right)\right\|_2^2\right],
\end{equation}
where $z_t$ represents the noised latent representation at time step $t$.
During inference, the reverse process starts with a random noise $x_T \sim \mathcal{N}(0, \mathbf{I})$ and iteratively generates an image sample from the noise from step $T$ to $0$. Stable diffusion also supports flexible conditional image generation through a cross-attention mechanism~\cite{46201}, enabling models to conditionally learn with various input modalities, such as text, semantic map, etc. Diffusion models are often used to generate additional training data with diverse styles to improve DA and DG performances.

\noindent\textbf{Segment Anything Model (SAM)}~\cite{kirillov2023segment} is a foundation model trained for promptable segmentation tasks, capable of producing high-quality masks for diverse segmentation prompts, including points, boxes, text, or masks. SAM consists of an image encoder for extracting image embeddings, a prompt encoder for embedding both sparse (points, boxes, text) and dense (masks) prompts, and a fast mask decoder that efficiently maps image and prompt embeddings to output masks. Trained on over $1$ billion masks, SAM demonstrates strong zero-shot segmentation performance. SAM is usually used to generate fine-grained instance-level masks for the refinement of predictions in DA and DG.

\subsection{Data Augmentation}
\begin{figure}[t]
  \centering
  \includegraphics[width=0.85\linewidth]{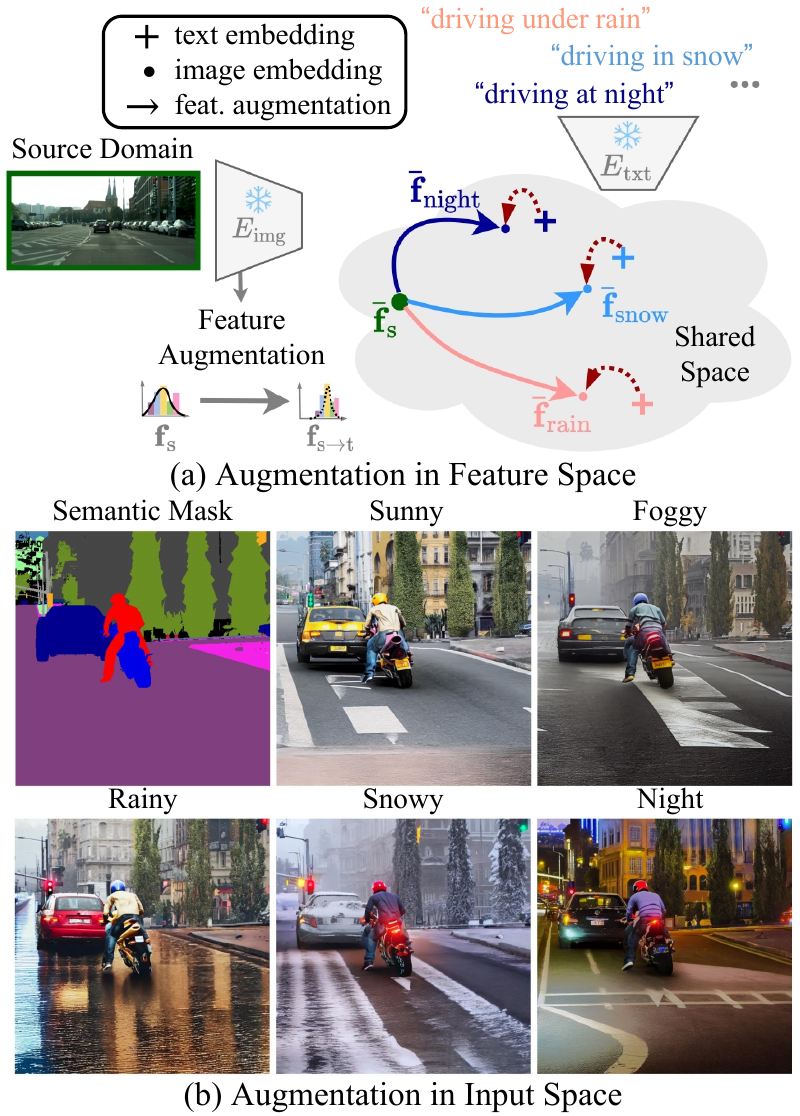}
   \vspace{-0.4cm}
   \caption{Examples of data augmentation using MFMs in feature space~\cite{fahes2023poda} and input space~\cite{jia2025dginstyle}.}
   \label{fig:aug}
\end{figure}

Several studies have leveraged MFMs to generate additional training data for augmentation, either in the feature space or input space (\cref{fig:aug}), to enhance generalization capabilities.

\subsubsection{Augmentation in Feature Space}
Generating data in feature space is more computationally efficient compared to directly generating images. For example, Dunlap et al.~\cite{dunlap2023using} learn a transformation of image embeddings from the training domain to unseen test domains using text descriptions. A classifier is then trained on both real and augmented embeddings, with domain alignment and class consistency losses ensuring that augmented embeddings remain in the correct domain and retain their class identity.
Similarly, Fahes et al.~\cite{fahes2023poda} employ general language descriptions of target domains to optimize affine transformations of source features, steering them towards target text embeddings while preserving semantic content. Given a source feature $\mathbf{f}_\text{s}$, they propose to generate stylized target feature $\mathbf{f}_{\text{s}\rightarrow\text{t}}$ by: 
\begin{equation}
	\mathbf{f}_{\text{s}\rightarrow\text{t}} = \bs{\sigma} \left( \frac{\mathbf{f}_\text{s} - \mu(\mathbf{f}_\text{s})}{\sigma(\mathbf{f}_\text{s})} \right) + \bs{\mu},
	\label{eqn:PIN}
\end{equation}
where $\mu(\cdot)$ and $\sigma(\cdot)$ are two functions returning channel-wise mean and standard deviation of input feature, ${\bs{\mu}}$ and~${\bs{\sigma}}$ are optimizable variables for target style driven by a prompt, which is the description embedding $\promptFeat$ of target domain (e.g., "driving under rain"):
\begin{equation}
\mc{L}_{\bs{\mu}, \bs{\sigma}}({\mb{f}}_{\text{s}\rightarrow\text{t}}, \promptFeat) = 1-\frac{{\mb{f}}_{\text{s}\rightarrow\text{t}}\cdot\promptFeat}{\|{\mb{f}}_{\text{s}\rightarrow\text{t}}\|\,\|\promptFeat\|}\,.
\label{eqn:loss_dir}
\end{equation}
%The augmented features are used to fine-tune segmentation models, enabling zero-shot domain adaptation. %PODA demonstrates efficacy across tasks such as semantic segmentation, object detection, and image classification.
Yang et al.~\cite{yang2024unified} extend the idea in~\cite{fahes2023poda} and achieve adaptation to diverse target domains without explicit domain knowledge. %domain-agnostic adaptation through Hierarchical Context Alignment, Domain-Consistent Representation Learning, and a Text-Driven Rectifier. These components align features, maintain semantic correlations, and rectify biases, enabling superior generalization without requiring domain identifiers or additional inference costs.
Cho et al.~\cite{cho2023promptstyler} simulate distribution shifts in the joint feature space by synthesizing diverse styles via prompts, eliminating the need for real images. %They train a linear classifier on synthesized style-content features to address source-free domain generalization, learning diverse style word vectors that preserve content information.
Recently, Vidit et al.~\cite{vidit2023clip} estimate a set of semantic augmentations using textual domain prompts and source domain images to transform source image embeddings into the target domain specified by the prompts.

\subsubsection{Augmentation in Input Space}
Some studies generate data directly in the image space to simulate domain shifts. For instance, Jia et al.~\cite{jia2025dginstyle} employ latent diffusion models~\cite{rombach2022high} to synthesize a diverse dataset of street scenes, which is then used to train a domain-agnostic semantic segmentation model. Similarly, Hemati et al.~\cite{hemati2024leveraging} leverage  a  diffusion-based framework for data-centric augmentation, enhancing  domain generalization by diversifying the training data.
Expanding on this approach, Singha et al.~\cite{singha2024unknown} generate proxy images for unknown classes using stable diffusion models, incorporating class-discriminative knowledge into visual embeddings.
In contrast, Benigmim et al.~\cite{benigmim2024collaborating} take a multi-model approach to domain-generalized semantic segmentation by integrating several foundation models. They leverage  CLIP for robust feature representation, diffusion models for generating synthetic images to enhance content diversity, and SAM for iterative prediction refinement.

\noindent\textbf{Summary and Insights.}
These approaches demonstrate the potential of MFMs in augmenting data across both feature and input spaces, offering effective means to improve model robustness under domain shifts. While feature-space augmentation is computationally efficient and avoids the pitfalls of image generation artifacts, input-space augmentation provides richer diversity in visual variations. However, challenges remain in ensuring that the augmented data effectively captures realistic target distributions without introducing harmful biases or overfitting to synthetic patterns. Future research should focus on adaptive augmentation strategies that dynamically tailor synthetic data generation to specific domain shifts, balancing efficiency and realism to maximize generalization benefits.

\subsection{Knowledge Distillation}
Distilling the rich knowledge of MFMs into smaller, more efficient models has been a key approach for enhancing domain generalization across various tasks. 
For instance, Zara et al.~\cite{zara2023unreasonable} integrate visual representations from pre-trained VLMs with source model knowledge and target data for source-free video domain adaptation. They leverage  CLIP's extensive domain knowledge for pseudo-labeling and knowledge distillation.
Similarly, Huang et al.~\cite{huang2023sentence} distill CLIP's semantic knowledge into a smaller student model, guiding its  representations to better align with CLIP’s text embeddings, which are both compact  and domain-invariant.
Building on this concept, Chen et al.~\cite{chen2024practicaldg} extend  knowledge distillation to open-set domain generalization by introducing  perturbations at the score, class, and instance levels, enabling lightweight vision models to inherit knowledge from large-scale VLMs.
Likewise, Tang et al.~\cite{tang2024source} tailor  VLMs for source-free domain adaptation through unsupervised prompt learning, embedding task-specific information, and distilling the adapted  VLM's knowledge into a target model.
Addepalli et al.~\cite{addepalli2024leveraging} focus on aligning the vision and language modalities of a teacher model with the vision modality of a pre-trained student model. They further refine this process by proposing an Align-Distill-Predict loss as:
\begin{equation}
\label{eq:vl2v-adip}
    \mathcal{L}_{\mathrm{ADiP}} = - \frac{1}{2n} \sum_{i=1}^n (\cos(\textbf{PF}^s_{x_i}, \textbf{T}_{y_i}) + \cos(\textbf{PF}^s_{x_i}, \textbf{I}^t_{x_i})),
\end{equation}
where $\textbf{PF}^s_{x_i}$ represents the projected features of the student model for the input image $x_i$, $\textbf{T}_{y_i}$ represents the text embedding corresponding to the ground truth class $y_i$, and~$\textbf{I}^t_{x_i}$ represent the image embeddings of the teacher model.
Similarly, Li et al.~\cite{li2023distilling} improve lightweight students' generalization by distilling fine-grained visual representations, enhancing vision-language alignment, and enriching teacher models with detailed semantic attributes.
Mistretta et al.~\cite{mistretta2025improving} advance prompt learning in VLMs by distilling knowledge from powerful models without requiring labeled data.

\noindent\textbf{Summary and Insights.} These works highlight the growing importance of knowledge distillation in adapting MFMs for domain generalization, offering a balance between efficiency and performance. While distillation enables lightweight models to inherit rich semantic knowledge, challenges remain in effectively transferring nuanced domain-specific information without excessive loss of generalization capabilities. Future research should focus on refining distillation techniques to better preserve task-relevant knowledge, explore adaptive distillation strategies for varying domain shifts, and investigate more efficient student architectures that can fully leverage the power of MFMs while maintaining computational efficiency.

\subsection{Learning Strategies}
In addition to data augmentation and knowledge distillation, novel  learning strategies are being developed to enhance  domain adaptation and generalization with MFMs.

\subsubsection{Prompt-based Strategies}
Prompt-based methods aim to optimize prompts for MFMs to improve DA and DG. For example, Zhang et al.~\cite{zhang2023domain} extend CLIP for DG by training a lightweight prompt generator that extracts  domain-specific cues  from input images and appends them to label prompts, effectively adapting CLIP’s representations to diverse domains.
Likewise, Ge et al.~\cite{ge2023domain} encode domain-specific information into prompts shared by images within  the same domain, enabling the classifier to dynamically adapt for different  domains.
Similarly, Chen et al.~\cite{chen2023multi} address multi-source DA by learning individual prompts for each source-target domain pair and mining relationships among these prompts to derive a shared, domain-invariant embedding space.
Building on this, Cheng et al. \cite{cheng2024disentangled} introduce a prompt-tuning framework that integrates  LLM-assisted text prompt disentanglement with text-guided visual representation disentanglement. . Their approach further incorporates domain-specific prototype learning, effectively balancing domain-specific and domain-invariant information.
Wang et al.~\cite{wang2024transitive} refine this strategy by employing vision prompts  to ensure domain invariance and language prompts to enhance class separability, dynamically balancing the two through an adaptive weighting mechanisms.
More recently, Xiao et al.~\cite{xiao2024any} propose  any-shift prompting, a probabilistic inference framework leveraging  hierarchical architecture and transformer inference networks to construct test prompts based on  training-test distribution relationships.
Additionally, Li et al.~\cite{li2024prompt} develop a dynamic object-centric perception network that uses prompt learning with object-centric gating and dynamic selective modules, enabling the model to focus on the most relevant spatial and channel features.
Lastly, Bai et al.~\cite{bai2025soft} introduce  a generative prompt-learning approach for DG, where domain-specific soft prompts are first trained, and instance-specific prompts are dynamically generated to adapt to  unseen target domains.

\subsubsection{Prior-based Strategies}
Prior-based methods leverage the strong domain knowledge   embedded in MFMs to improve performance in domain adaptation and generalization. For instance, Peng et al.~\cite{peng2025learning} address domain adaptation for 3D segmentation by leveraging SAM~\cite{kirillov2023segment} to incorporate 2D prior knowledge, facilitating the alignment of features across diverse 3D data domains into a unified representation.
Similarly, Lim et al.~\cite{lim2024cross} integrate the rich semantic knowledge of VLMs with segment reasoning from traditional domain adaptation methods to relabel novel classes in the target domain. This approach enables effective adaptation to extended taxonomies without requiring ground truth labels in the target domain.
Xu et al.~\cite{xu2024visual} enhance cross-modal unsupervised domain adaptation by utilizing the prior knowledge encoded in vision foundation models to generate more accurate labels for unlabeled target domains.
Taking a different approach, Zhu et al.~\cite{zhu2024clip} employ CLIP to quantify domain divergence through domain-agnostic distributions and calibrate target pseudo-labels with language guidance, effectively reducing the domain gap. 
A recent work by Sun et al.~\cite{sun2025dynalign} integrates DA with foundation models to bridge both image-level and label-level domain gaps by leveraging prior semantic knowledge for category alignment, dynamically adapting to new taxonomies, and enabling fine-grained segmentation without manual annotations.
Additionally, several works~\cite{prabhudesai2023diffusion,du2023diffusion,gao2023back} leverage diffusion priors for efficient adaptation, leveraging the generative capabilities of diffusion models to improve feature alignment and robustness across domains. 

\subsubsection{Refinement-based Strategies}
Refinement-based methods leverage  MFMs to enhance  feature representations  or pseudo-labels, improving adaptation and generalization across domains. 
%Wei et al.~\cite{wei2024stronger} introduce a robust, parameter-efficient fine-tuning approach  for utilizing VFMs in domain-generalized semantic segmentation. Their method  incorporates learnable tokens tied to instances,  enabling feature refinement at the instance level within each model layer.
For example, Lai et al.~\cite{lai2024empowering} propose an efficient adaptation strategy for VLMs that maintains  their original knowledge while maximizing flexibility for learning new information. They further  design a domain-aware pseudo-labeling scheme specifically tailored for VLMs, facilitating effective domain disentanglement.
Taking  a complementary approach, Hu et al.~\cite{hu2024reclip} mitigate misaligned visual-text embeddings by learning a projection space and generating pseudo-labels. They iteratively apply cross-modality self-training to update visual and text encoders, refine labels, and reduce domain gaps and misalignments.
Recently, Xia et al.~\cite{xia2024unsupervised} leverage text-to-image diffusion models pre-trained on large-scale image-text datasets to enhance cross-modal capabilities for semantic segmentation. They achieve state-of-the-art performance across various modality tasks, including images-to-depth, infrared, and event modalities.
Zhang et al.~\cite{zhang2024source}  integrate CLIP's vision encoder with its zero-shot text-based classifier, refining the fitted classifier for source-free DA.
%These works highlight the growing role of MFMs in addressing DA and DG challenges, showcasing their versatility in both feature refinement and pseudo-labeling.

\subsubsection{Other Strategies}
Beyond the previously discussed approaches, additional  learning strategies have been developed for domain adaptation and generalization with MFMs. For instance, Cha et al.~\cite{cha2022domain} introduce mutual information regularization with Oracle, robustly approximating Oracle models using large pre-trained models like CLIP~\cite{radford2021learning}.
Shu et al.~\cite{shu2023clipood} adapt CLIP to handle domain shifts and open-class scenarios by incorporating  margin metric softmax to capture semantic relationships between text classes. They further enhance optimization by combining a zero-shot model with a fine-tuned task-adaptive model using the beta moving average.
Addressing a different challenge, Lai et al.~\cite{lai2023padclip} tackle  catastrophic forgetting when fine-tuning CLIP on target domains by introducing catastrophic forgetting measurement that dynamically adjusts the learning rate. %It extends DebiasPL~\cite{wang2022debiased} for pseudo-labeling across multiple domains.
Zara et al.~\cite{zara2023autolabel} propose an object-centric compositional labeling approach for target-private classes, improving  the rejection of target-private instances while enhancing  alignment between shared classes across domains.
Yu et al.~\cite{yu2024open} introduce an entropy optimization strategy to facilitate  open-set DA by leveraging  CLIP outputs.
More recently, Li et al.~\cite{li2024split} introduce a disentanglement approach that separates CLIP representations  into language- and vision-specific components. Their method utilizes modality-ensemble training to balance modality-specific nuances and shared information while integrating a modality discriminator for improved cross-domain alignment.

\noindent\textbf{Summary and Insights.} Prompt-based methods offer a lightweight and efficient way to adapt MFMs with minimal fine-tuning, whereas prior-based approaches leverage the rich domain knowledge embedded in MFMs to guide adaptation. Refinement-based methods improve feature representations and pseudo-labels, ensuring more reliable adaptation. Meanwhile, emerging strategies such as mutual information regularization and hybrid zero-shot and fine-tuned approaches push the boundaries of DA and DG.
Despite these advancements, challenges remain in efficiently adapting MFMs to highly diverse domains while preserving their generalization ability. Future research should explore how to effectively combine multiple strategies—such as integrating prompt-based tuning with prior-based learning—to maximize both adaptability and computational efficiency. Additionally, understanding the theoretical underpinnings of these strategies could provide deeper insights into their limitations and optimal use cases. Developing more robust benchmarks to evaluate the effectiveness of different strategies across real-world domain shifts will also be crucial in advancing this field.

% Zhang et al.~\cite{zhang2025learning} integrate a Vision Transformer (ViT)-based foundation model into the stereo matching pipeline, improving the zero-shot performance of stereo matching networks on unseen domains.
% Yu et al.~\cite{yuclipceil} enhance CLIP's zero-shot DG by refining visual feature channels to ensure domain invariance and class relevance, maintaining image-text alignment, and employing a self-attention fusion module to integrate multi-scale features.
% Kukleva et al.~\cite{kukleva2024x} align frozen text embeddings to videos within a shared embedding space through an adapter architecture that disentangles learnable temporal modeling from the frozen visual encoder.
% Vesdapunt et al.~\cite{vesdapunt2025hvclip} addresses catastrophic forgetting in DA by mapping CLIP features to a high-dimensional hypervector space. This approach enhances robustness through discrepancy reduction to mitigate domain shifts and employs feature augmentation to synthesize labeled target features.

\section{Adaptation of Multimodal Foundation Models}
\label{sec:adaf}
\begin{figure}[t]
  \centering
  \includegraphics[width=0.9\linewidth]{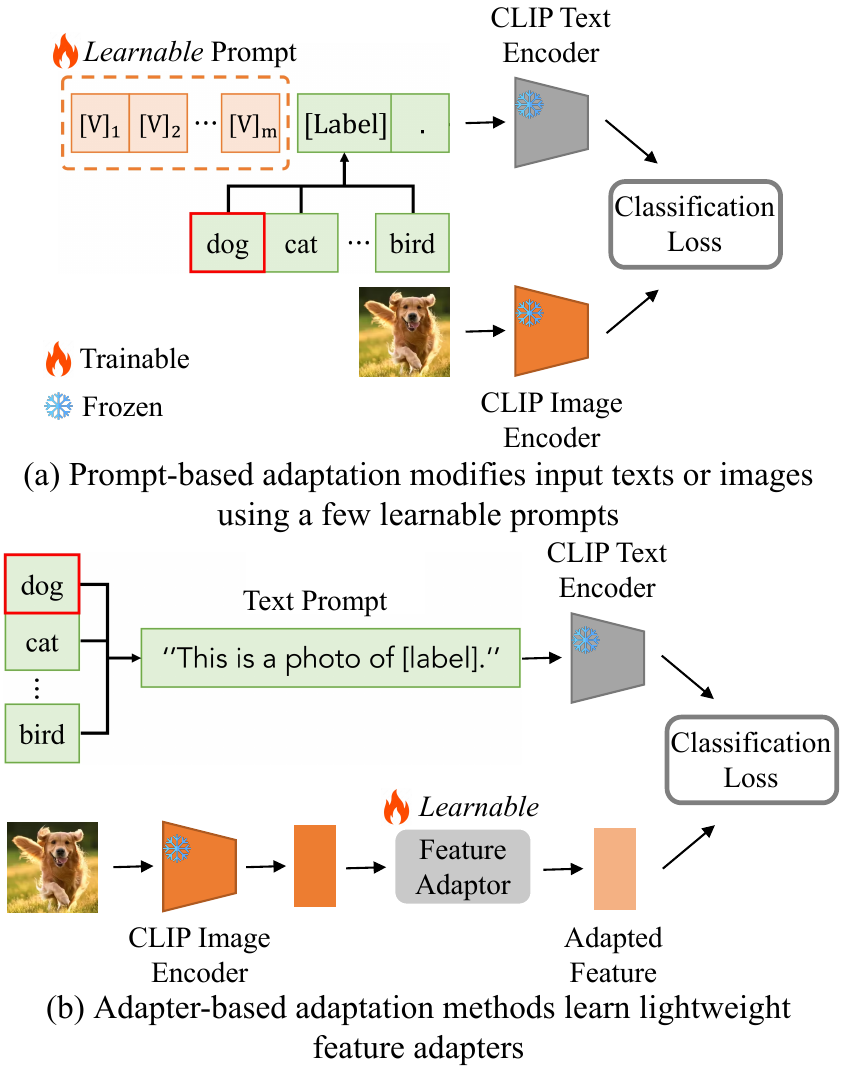}
   \vspace{-0.5cm}
   \caption{A comparison of prompt-based and adapter-based adaptation.}
   \label{fig:ada}
\end{figure}

Unlike previous approaches that leverage MFMs for domain adaptation and generalization tasks, this section focuses on adapting the MFMs themselves to better handle distribution shifts in downstream tasks. Various transfer learning strategies, such as prompt tuning and feature adapters, have been developed to effectively adapt MFMs to downstream tasks. \cref{fig:ada} illustrates the difference between prompt-based and adapter-based adaptation.%\cref{fig-afm} provides an overview of representative methods, and \cref{fig:ada} illustrates the difference between prompt-based and adapter-based adaptation.

% \begin{figure}[t!]
% 	\centering
% 	\resizebox{.5\textwidth}{!}{
% 	\input{imgs/fig-AFM}
% 	}
%    \vspace{-0.7cm}
% 	\caption{Taxonomy of Adaptation of MFMs.}
% 	\label{fig-afm}
% \end{figure}

\subsection{Prompt-based Adaptation}
Prompt-based adaptation modifies input texts or images using a few learnable prompts for parameter-efficient tuning, avoiding the need to fine-tune the entire model.

\subsubsection{Text Prompt Tuning}
Text prompt tuning methods refine  the input to the text encoder,  enabling  MFMs to better adapt to downstream tasks.
For example, CoOp~\cite{zhou2022learning} adapts CLIP-like VLMs for image recognition by treating  a prompt’s context words as learnable vectors, inspired by prompt learning in NLP~\cite{liu2023pre}. Rather than relying on fixed templates like “a photo of a [Label]”, CoOp represents prompts as "$[V]_1$$[V]_2$...$[V]_M$[Label]", where each $[V]_m$ is a trainable vector matching the word embedding dimension, optimized using cross-entropy loss.
To mitigate CoOp’s tendency to overfit to base classes, CoCoOp~\cite{zhou2022conditional} introduces input-conditional tokens for each image. In addition to the $M$ learnable context vectors, CoCoOp trains  a lightweight neural network $h_{{\theta}}(\cdot)$ to dynamically modify the context based on image features, updating each token as  $[V]_m ({x}) = [V]_m +  h_{{\theta}} ({x})$ and $m \in \{1, 2, ..., M\}$.
Further addressing overfitting, Ma et al.~\cite{ma2022understanding} propose a gradient projection  technique that restricts updates to a low-rank subspace defined by early-stage gradient flow eigenvectors.
Bulat et al.~\cite{bulat2022language} take a different approach by constraining  learned prompts to remain close to handcrafted ones, reducing base class overfitting.
Lu et al.~\cite{lu2022prompt} shift the focus  from learning input embeddings to modeling  the distribution of output embeddings,  providing a more stable adaptation mechanism.
In a similar vein, Derakhshani et al.~\cite{derakhshani2022variational} adopt a probabilistic approach, modeling the input prompt space  as a prior distribution, while Zhu et al.~\cite{zhu2022prompt} selectively update prompts whose gradients align with general knowledge, preventing knowledge forgetting.
He et al.~\cite{he2022cpl} introduce  counterfactual generation and contrastive learning to enhance tuning.
A novel direction is explored  by Chen et al.~\cite{chen2022prompt}, who employ optimal transport to learn diverse prompts, capturing the comprehensive characteristics of multiple  categories and improving generalization across domains.

Additionally, Sun et al.~\cite{sundualcoop}  introduce a dual-prompt learning strategy, where a pair of positive and negative prompts is trained to enhance multi-label recognition. Guo et al.~\cite{guo2022texts} take a different approach by leveraging text descriptions to guide  prompt learning, extracting both coarse- and fine-grained embeddings to improve multi-label classification  performance.
Ding et al.~\cite{ding2022prompt} extend prompt tuning to  a multi-task learning  by employing a task-shared meta-network that generates task-specific prompt contexts, enabling efficient adaptation across different domains.
 To enhance generalization to unseen classes, Yao et al.~\cite{yao2023visual} propose a method that  minimizes the discrepancy between learnable and handcrafted prompts, ensuring better alignment with pre-trained representations.
More recently, Wu et al.~\cite{wu2024protect} improved prompt tuning for open-set classification by introducing a hierarchical calibration mechanism that aligns predictions across semantic label hierarchies.

% UPL~\cite{huang2022unsupervised} introduces unsupervised learning to prompt tuning, removing the dependency on labeled data. 
% AAPE~\cite{huang2024aggregate} distills textual knowledge from natural language prompts into prompt embeddings, aligning them with input images to optimize downstream task performance.

\subsubsection{Visual Prompt Tuning}
Visual prompt tuning methods modify the input to the image encoder input to adapt MFMs, enabling improved task-specific performance without modifying the backbone model.
For example, Bahng et al.~\cite{bahng2022exploring} adapt CLIP by learning a single task-specific image perturbation that guides the frozen model toward  new tasks, achieving  performance comparable to linear probes. They aim to learn a single, task-specific visual prompt $v_\theta$ parameterized by $\theta$. The prompt is added to the input image to form a prompted image $x+v_\theta$. The model maximizes the likelihood of the correct label $y$ to update $\theta$ during training. During evaluation, the optimized prompt is added to all test-time images, effectively steering model predictions.
Rong et al.~\cite{rong2023retrieval} enhance fine-grained classification in few-shot learning by incorporating a retrieval mechanism that references relevant samples during inference, improving the model's ability to generalize with limited data. 
Taking a different approach, Wu et al.~\cite{wu2022unleashing} introduce  a learnable prompt wrapped around a resized image, leveraging input diversity and gradient normalization techniques. to improve adaptation. This strategy ensures that the model effectively captures domain-specific nuances while maintaining the generalization ability of the pre-trained image encoder.

\subsubsection{Text-Visual Prompt Tuning}
Text-visual prompt tuning methods modify both text and visual encoder inputs to adapt MFMs.
For example, Zang et al.~\cite{zang2022unified} combine text and visual prompt tuning by optimizing prompts across both modalities using a small neural network, achieving improved trade-offs in few-shot learning and domain generalization compared to unimodal approaches.
Shen et al.~\cite{shen2022multitask} leverage cross-task knowledge to enhance prompt tuning, learning transferable prompts from multiple source tasks and enabling joint adaptation across tasks.
Khattak et al.~\cite{khattak2022maple} jointly learn dynamic prompts for both vision and language branches, ensuring strong coupling and stage-wise feature alignment for improved generalization.
Similarly, Xing et al.~\cite{xing2023dual} improve prompt tuning by jointly learning text and visual prompts, incorporating a class-aware visual prompt tuning scheme that dynamically generates visual prompts through cross-attention between text prompts and image features. 
To enhance robustness, Abdul et al.~\cite{abdul2024align} align test sample statistics with source data through prompt tuning, while Khattak et al.~\cite{khattak2023self} propose  a self-regularization framework for CLIP prompt learning, guiding prompts to optimize both task-specific and task-agnostic representations through mutual agreement.
Recent advances further refine these strategies. Wu et al.~\cite{wu2025cascade} introduce a two-phase framework, first extracting domain-general knowledge from a larger teacher model using boosting prompts, followed by task-specific prompt adaptation.
 Hao et al.~\cite{hao2025quantized} address overfitting and catastrophic forgetting by applying quantization as a lightweight regularization technique.

\noindent\textbf{Summary and Insights.} Prompt-based adaptation has emerged as a powerful and efficient alternative to traditional fine-tuning, leveraging learnable text and visual prompts to guide  MFMs toward specific tasks while preserving their broad generalization capabilities. Although  prompt-based tuning has significantly improved parameter efficiency and adaptability, several critical challenges  remain. How can we design prompts that generalize across multiple domains without task-specific tuning? Can we develop self-supervised or unsupervised prompt learning methods to eliminate  dependence on labeled adaptation data? Addressing these challenges will be essential  to fully  unlocking the potential of prompt-based adaptation, enabling more flexible, scalable, and autonomous learning for foundation models across diverse applications.

\subsection{Adapter-based Adaptation}
Adapter-based adaptation methods leverage lightweight feature adapters to fine-tune MFMs.
For example, Clip-Adapter~\cite{gao2021clip} enhances VLMs by fine-tuning feature adapters on either the visual or language branch. It employs a bottleneck layer and residual-style feature blending, outperforming context optimization while maintaining a simple design. The adapter is a two-layer MLP with parameters $W_1, b_1, W_2, b_2$. Given an input image feature $f_c$, the adapted feature $f_a$ is:
\begin{align}
\begin{split}
    f_a = \varphi(f_c W_1^T + b_1) W_2^T + b_2,
\end{split}
\end{align}
where $\varphi$ denotes the activation function in the MLP. The adapted feature $f_a$ is linearly combined with the $f_c$ with a hyper-parameter $\alpha \in [0,1]$ to output the final prediction:
\begin{align}
\label{v1}
\begin{split}
    \hat{p} = \alpha f_a W_c^T + f_c W_c^T,
\end{split}
\end{align}
where $W_c$ is the weight of the text classifier. The parameters of MLP are optimized by minimizing the cross-entropy loss between predictions and ground truth labels.
Tip-Adapter~\cite{zhang2021tip} improves few-shot classification by generating adapter weights through a key-value cache model from the few-shot training set. This approach eliminates the need for backpropagation and training, achieving performance comparable to or exceeding that of Clip-Adapter.
SVL-Adapter~\cite{pantazis2022svl} combines vision-language pretraining with self-supervised representation learning to enhance low-shot image classification.
Additionally, Kahana et al.~\cite{kahana2022improving} adapt CLIP for both regression and classification tasks on unlabeled datasets by incorporating a distributional prior over labels. Their approach trains an adapter network to minimize prediction variability  while ensuring alignment with the prior distribution, enhancing model reliability in label-scarce scenarios. 
%SgVA-CLIP~\cite{peng2022sgva}  improves few-shot image classification by employing implicit knowledge distillation, a vision-specific contrastive loss, and a cross-modal contrastive loss to produce more discriminative visual features.
Karmanov et al.~\cite{karmanov2024efficient} propose  a training-free dynamic adapter for TTA with VLMs. Their approach leverages a lightweight key-value cache and progressive pseudo-label refinement, enabling  efficient  adaptation to test data without backpropagation. To further mitigate label noise, they  incorporate negative pseudo-labeling, improving robustness in dynamic environments.
Lu et al.~\cite{lu2024improving} address  prediction bias in VLMs by introducing  variational adapters with learnable textual tokens. These adapters effectively  separate base and novel classes in latent space, refining the model's ability to generalize to unseen categories.
Additionally, SAM-Adapter~\cite{chen2023sam}, MA-SAM~\cite{chen2024ma}, and CAT-SAM~\cite{xiao2025cat} enhance the SAM by integrating  domain-specific information through lightweight  adapters, improving segmentation performance in specialized tasks. 
%MA-SAM~\cite{chen2024ma} adapts SAM for medical image segmentation by efficiently fine-tuning it with 3D adapters inserted into the image encoder to incorporate volumetric/temporal information while preserving SAM's pre-trained 2D backbone.

\noindent\textbf{Summary and Insights.} Adapter-based adaptation has emerged as a promising approach for efficiently fine-tuning MFMs while preserving their generalization ability. While adapters provide efficiency and flexibility, their effectiveness relies heavily on the quality of feature representations extracted by the frozen backbone. When  the pre-trained model encounters  out-of-distribution data, adapters alone may not be sufficient for robust adaptation. Moreover, integrating adapter-based techniques with foundation models like SAM highlights the potential to extend adaptation beyond classification to structured tasks such as segmentation.

\subsection{Other Adaptation Methods}
\subsubsection{Fine-tuning Methods}
Some approaches  fine-tune all parameters of MFMs. For example, Wortsman et al.~\cite{wortsman2022robust} enhance robustness to distribution shifts in large VLMs by ensembling the weights of zero-shot and fine-tuned models. Li et al.~\cite{li2022masked} improve CLIP through unsupervised fine-tuning on unlabeled target domain data, leveraging  pseudo-labeling and regularization to jointly optimize global and local features.

\subsubsection{Training-free Methods}
Another promising direction focuses on adapting MFMs without any parameter tuning. For instance, Udandarao et al.~\cite{udandarao2022sus} enhance VLMs by constructing dynamic support sets—either via image generation or retrieval—and leveraging image-text distances to refine classification without additional training.
Guo et al.~\cite{guo2022calip} enhance CLIP's zero-shot performance using a parameter-free cross-modal attention module that adaptively aligns visual and textual features, eliminating the need for additional training or learnable parameters. 
Zanella et al.~\cite{zanella2024test} improve zero-shot and few-shot VLM performance by jointly optimizing view quality assessment and density mode seeking. 
Zhang et al.~\cite{zhang2024dual2} introduce dual memory networks, where  static memory caches  training data knowledge and dynamic memory adapts  online test features. %This training-free approach effectively addresses zero-shot, few-shot, and training-free few-shot adaptation scenarios.
Recently, Wang et al.~\cite{wang2024hard} applied  Gaussian assumptions for class features, enabling training-free integration of visual and textual modalities, while Ge et al.~\cite{ge2023improving} improve accuracy by identifying ambiguous predictions through prompt and transformation consistency, augmenting text prompts with semantic labels from the WordNet hierarchy.

\subsubsection{LLM-based Methods}
Large language models (LLMs) also aid in adapting VLMs. For instance, Pratt et al.~\cite{pratt2022does} enhance  open-vocabulary image classification  generating discriminative prompts with LLMs, improving accuracy without additional training or task-specific knowledge.
Menon et al.~\cite{menon2022visual} enhance VLM-based classification by querying LLMs  for descriptive features.
More recently, Parashar et al.~\cite{parashar2024neglected} leverage LLMs to identify frequent concept synonyms in pretraining data, enabling  more effective prompting.

\subsubsection{Test-Time Adaptation of VLMs}
Recent works also explore test time adaptation of VLMs. 
For example, Shu et al.~\cite{shutest} dynamically adjust prompts for each test sample by minimizing entropy across augmented views. Feng et al.~\cite{feng2023diverse} take a different approach, using pre-trained diffusion models for diverse data augmentation and cosine similarity-based filtration to enhance test-time prompt fidelity.
Ma et al.~\cite{ma2024swapprompt} introduce a self-supervised contrastive learning framework with dual prompts—an online prompt and a historical target prompt—combined with a swapped prediction mechanism for improved adaptation.
%BoostAdapter~\cite{zhangboostadapter} bridges training-required and training-free approaches by leveraging a lightweight key-value memory for feature retrieval from historical target distribution samples and regionally bootstrapped test samples, enabling efficient and effective adaptation of CLIP to diverse downstream tasks.
% RLCF~\cite{zhao2023test}  enhances zero-shot generalization in VLMs by using CLIP as a reward model during test-time adaptation to correct predictions.
Osowiechi et al.~\cite{osowiechi2024watt} improve prediction accuracy by leveraging weight averaging with different text prompts and incorporating text embedding averaging.
Similarly, Farina et al.~\cite{farina2024frustratingly} improve  generalization by augmenting predictions, retaining only the most confident ones, and marginalizing them using a zero Softmax temperature -- all without backpropagation. 
% DPE~\cite{zhang2024dual} progressively accumulates task-specific knowledge from both textual and visual prototypes, optimizing multimodal representations with learnable residuals. 

\subsubsection{Dense Prediction Tasks}
VLMs can also be adapted  for dense prediction tasks such as semantic segmentation. Rao et al.~\cite{rao2022denseclip} introduce a frameworkthat repurposes pre-trained CLIP knowledge by transforming image-text matching into pixel-text matching, and leveraging  pixel-text score maps with contextual prompts for guidance.
Similarly, Zhou et al.~\cite{zhou2022extract} enhance  pixel-level dense prediction by integrating  CLIP embeddings with pseudo-labeling and self-training.
More recently, Zhang et al.~\cite{zhang2024improving} improve the robustness and efficiency of SAM for image segmentation under significant distribution shifts by introducing a weakly supervised self-training strategy with anchor regularization and low-rank fine-tuning.

\subsubsection{Other Methods}
Several other methods have been proposed for adapting MFMs. For instance, Zhang et al.~\cite{zhang2021vt} improve CLIP’s transfer performance by leveraging  visual-guided text features that adaptively explore informative image regions and aggregate visual features through attention, improving  semantic alignment for downstream classification tasks.
Yu et al.~\cite{yu2023task} refine  VLMs transfer by tuning a residual to the pre-trained text-based classifier, preserving prior knowledge while enabling task-specific adaptation.
Ouali et al.~\cite{ouali2023black} propose a computationally efficient black-box method for vision-language few-shot adaptation, operating on pre-computed features and aligning image-text representations using  a closed-form least-squares initialization with a re-ranking loss.
 Zanella et al.~\cite{zanella2024low} introduce low-rank adaptation~\cite{hu2021lora} for few-shot learning in VLMs, achieving significant improvements with reduced training time and consistent hyperparameters across  datasets.
Xuan et al.~\cite{xuan2024adapting} enhance VLMs by extracting task-agnostic knowledge from the text encoder and injecting it into input image and text features.
More recently, Zhang et al.~\cite{zhang2024rethinking} address  data misalignment in VLMs by decoupling task-relevant and task-irrelevant knowledge using a structural causal model.
Lin et al.~\cite{lin2023multimodality} extend adaptation by learning from both visual and non-visual data (e.g., text and audio), repurposing class names as additional one-shot training samples to improve performance across vision, language, and audio tasks.
% Liang et al.~\cite{liang2024umfc} enhance the transferability of vision-language models by mitigating domain-specific biases in the visual and text encoders, making them domain-invariant without requiring labeled data or additional optimization. 
% Zang et al.~\cite{zang2024overcoming} improve VLMs generalization by synthesizing out-of-distribution (OOD) features using a class-conditional feature generator and regularizing decision boundaries through adaptive self-distillation during joint optimization.
% Zhu et al.~\cite{zhu2024croft} improve out-of-distribution generalization under covariate shifts by minimizing the gradient magnitude of energy scores and detecting semantic-shifted unseen classes.

\noindent\textbf{Summary and Insights.} The adaptation of MFMs is evolving along multiple promising directions, each balancing efficiency and effectiveness. Future research should explore hybrid approaches that integrate multiple  strategies—such as combining prompt tuning with lightweight adapters—to optimize both  efficiency and task-specific performance. Additionally, a deeper understanding of MFMs failure modes  under distribution shifts is essential for developing  more robust and generalizable adaptation techniques. Lastly, as adaptation techniques evolve, ensuring  stability and interpretability will be critical, especially in safety-critical applications  like  autonomous driving and medical imaging.

\section{Revisiting Traditional DA/DG Paradigms in the Era of Foundation Models}
\label{sec:Revisiting}

The emergence of large-scale MFMs, pre-trained on massive web datasets, fundamentally challenges the traditional paradigms of DA/DG. These older methods were built on the assumption of adapting models from a limited source domain to a distinct target domain. However, the vast and diverse training data of MFMs means they often handle distribution shifts implicitly, exhibiting powerful zero-shot and few-shot generalization without explicit adaptation. While MFMs have undoubtedly shifted the landscape, we argue that traditional DA/DG is not obsolete but is rather being repositioned and redefined. 

\subsection{The Fragility of MFMs}
Despite their impressive generalization capabilities, MFMs have significant fragilities where traditional DA/DG methods find renewed purpose. The zero-shot performance of MFMs often decreases on tasks requiring fine-grained distinctions~\cite{zhou2022learning}, such as identifying specific species, or abstract reasoning, like counting objects. This is because their training objective prioritizes global feature alignment over local details. Furthermore, MFMs can fail unexpectedly when presented with data that is truly out-of-distribution~\cite{mayilvahanan2025in} relative to their web-scale pre-training data (e.g., medical imaging, satellite imagery, industrial inspection, scientific data). These limitations underscore that even massive pre-training does not guarantee universal robustness, necessitating specialized adaptation techniques to correct for these specific failure modes. 

\subsection{The Repositioning of DA/DG in the MFM Era}
The focus of DA/DG is shifting in the MFM era. Instead of learning domain-invariant features from scratch, the new challenge is to efficiently adapt the rich knowledge already encoded within MFMs to a specific target application. This has given rise to a new class of adaptation methods that are intellectually descended from traditional DA/DG. Foundational ideas such as distribution alignment and adversarial learning in DA/DG have become the conceptual backbone for emerging parameter-efficient adaptation methods, including prompt tuning and adapters. In conclusion, while the assumption of a limited, well-defined training domain in traditional DA/DG is fundamentally challenged by the broad generality of MFMs, this does not render DA/DG obsolete. Rather, it redefines the field, opening up a new set of research directions focused on how to most effectively leverage, adapt, and specialize the powerful prior knowledge embedded in MFMs. The central challenge now lies in achieving this adaptation in a manner that is robust, efficient, and secure, ensuring that MFMs can meet the demands of diverse downstream tasks and real-world domains.

\begin{table*}[ht!]
    \centering
    %\tabstyle{5pt}
    \caption{Commonly used multimodal adaptation and generalization datasets. V: video, A: audio, F: optical flow, P: point cloud, I: image.}
   \vspace{-0.4cm}
    \label{tab:datasets}
    \resizebox{\textwidth}{!}{
    \begin{tabular}{l r c l}
    \toprule
   \textbf{Dataset} & \textbf{Modality} & \textbf{\#Domains} & \textbf{Characterization of domain shift} \\

    \textbf{Action recognition} & \\
    \quad- HAC~\cite{dong2023simmmdg} & V+A+F & 3 & Human, animal, and cartoon \\
    \quad- EPIC-Kitchens~\cite{munro2020multi} & V+A+F & 3 & Three different kitchens \\
    \quad- CharadesEgo~\cite{sigurdsson2018actor} & V+A & 2 & First and third-person view \\
    \quad- ActorShift~\cite{zhang2022audio} & V+A & 2 &  Human, animal \\
    \quad- UCF$\rightarrow$\{HMDB, Olympic\}~\cite{soomro2012ucf101,kuehne2011hmdb,niebles2010modeling} & V+F & 3 & Dataset to dataset (see~\cite{xiong2024modality}) \\
    % \quad- UCF-Olympic~\cite{soomro2012ucf101,niebles2010modeling} & V+F & 2 & Dataset to dataset (see~\cite{xiong2024modality}) \\
    \quad- Kinetics50-C~\cite{yang2023test} & V+A & - & Artificial corruptions \\
    \quad- Kinetics-100-C~\cite{dong2025aeo} & V+A & - & Artificial corruptions \\
    \midrule

    \textbf{Semantic segmentation} & \\
    \quad- nuScenes Day/Night~\cite{caesar2020nuscenes} & P+I & 2 & Day to Night (see~\cite{jaritz2020xmuda}) \\
    \quad- nuScenes USA/Singapore~\cite{caesar2020nuscenes} & P+I & 2 & Country to country (see~\cite{jaritz2020xmuda})  \\
    \quad- A2D2$\rightarrow$SemanticKITTI~\cite{geyer2020a2d2,behley2019semantickitti} & P+I & 2 & Dataset to dataset (see~\cite{jaritz2020xmuda}) \\
    \quad- SYNTHIA$\rightarrow$SemanticKITTI~\cite{ros2016synthia,behley2019semantickitti} &  P+I& 2 & Synthetic to real (see~\cite{shin2022mm}) \\
    \quad- SemanticKITTI$\rightarrow$SYNTHIA~\cite{ros2016synthia,behley2019semantickitti} &  P+I& 2 & Real to synthetic (see~\cite{cao2023multi}) \\
    \quad- SemanticKITTI$\rightarrow$Waymo~\cite{ros2016synthia,behley2019semantickitti} &  P+I& 2 & Dataset to dataset (see~\cite{cao2023multi}) \\
    \quad- GTA5$\rightarrow$\{Cityscapes, BDD100K, Mapillary\}~\cite{richter2016playing,cordts2016cityscapes,yu2020bdd100k,neuhold2017mapillary} & I & 4 & Synthetic to real (see~\cite{wei2024stronger}) \\
    \quad- Cityscapes$\rightarrow$\{BDD100K, Mapillary\}~\cite{cordts2016cityscapes,yu2020bdd100k,neuhold2017mapillary} & I & 3 & Dataset to dataset (see~\cite{wei2024stronger}) \\
    \midrule

    \textbf{Image classification} & \\
    \quad- VLCS~\cite{fang2013unbiased} & I & 4 & Dataset to dataset  \\
    \quad- OfficeHome~\cite{venkateswara2017deep} & I  & 4 & Art, clipart, product, real \\
    \quad- PACS~\cite{li2017deeper} & I & 4 & Photo, art, cartoon, sketch \\
    \quad- DomainNet~\cite{peng2019moment} & I  & 6 & Clipart, infograph, painting, quickdraw, real, sketch \\
    \quad- Wilds~\cite{koh2021wilds} & I  & - & Camera, hospital, batch, scaffold, location, time, etc \\
    % \quad- ImageNet-Sketch~\cite{wang2019learning} & I  & - & Sketch images \\
    % \quad- ImageNet-R~\cite{hendrycks2021many} & I  & - & Image style changes \\
    % \quad- ImageNet-A~\cite{hendrycks2021natural} & I  & - & Naturally adversarial examples \\

    \bottomrule
    \end{tabular}
    }
\end{table*}

\section{Datasets and Applications} 
\label{sec:application}

\begin{figure*}[t]
  \centering
  \includegraphics[width=0.95\linewidth]{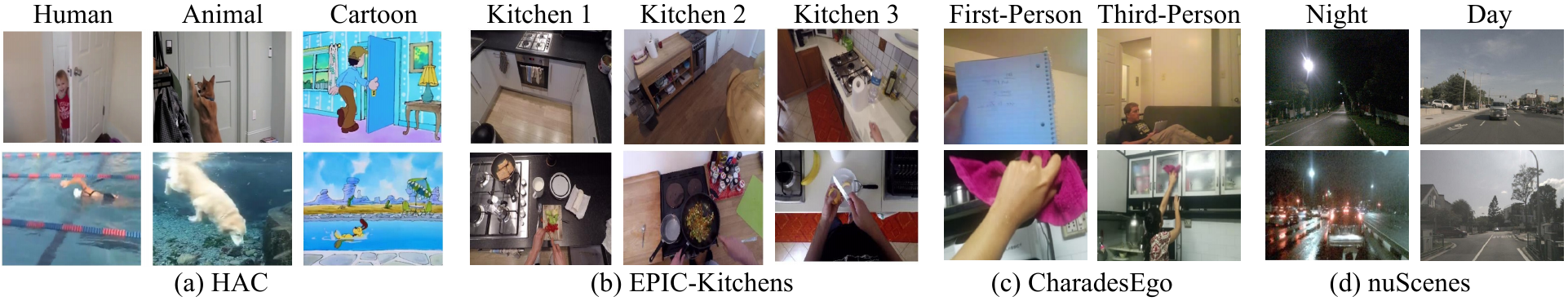}
   \vspace{-0.5cm}
   \caption{Examples from various datasets manifesting different types of domain shift. In (a), the domain shift mainly corresponds to the person, animals, or cartoon characters that perform actions. In (b), the domain shift is caused by styles in different kitchens. In (c), viewpoint changes are the main reason for the domain shift. In (d), the domain shift is caused by day and night changes.}
   \label{fig:dataset}
\end{figure*}

Multimodal adaptation and generalization have been explored  across various  application domains, including action recognition, semantic segmentation, image classification, sentiment analysis, person re-identification, and depth completion. Benchmark datasets play a crucial role in evaluating and comparing adaptation techniques, providing standardized testbeds to assess model performance under diverse conditions. An overview of commonly used datasets is shown in  \cref{tab:datasets}, while \cref{fig:dataset} illustrates  examples of different types of domain shifts. 

\subsection{Datasets for Action Recognition}
The EPIC-Kitchens~\cite{munro2020multi} and Human-Animal-Cartoon (HAC)~\cite{dong2023simmmdg} datasets are widely   for MMDA and MMDG tasks in  action recognition. EPIC-Kitchens includes $10,094$ video clips covering eight actions ("put", "take", "open", "close", "wash", "cut", "mix", and "pour"), recorded in three distinct kitchens, forming three separate domains D1, D2, and D3. The HAC dataset features seven actions ("sleeping", "watching tv", "eating", "drinking", "swimming", "running", and "opening door") performed by humans, animals, and cartoon figures, spanning three domains H, A, and C with $3,381$ video clips. Both datasets provide  video, optical flow, and audio modalities, making them valuable for studying multimodal adaptation. The Additionally, the CharadesEgo~\cite{sigurdsson2018actor} (domain shift from first and third-person view) and ActorShift~\cite{zhang2022audio} (domain shift from human and animal) datasets are also used in the  literature. For MMTTA, UCF~\cite{soomro2012ucf101}, HMDB~\cite{kuehne2011hmdb}, Olympic~\cite{niebles2010modeling}, and Kinetics-600~\cite{carreira2018short} with corruptions are used to simulate  different types of domain shifts.

\subsection{Datasets for Semantic Segmentation}
The nuScenes~\cite{caesar2020nuscenes}, A2D2~\cite{geyer2020a2d2}, and SemanticKITTI~\cite{behley2019semantickitti} datasets  are  among the most  commonly used benchmarks for 3D semantic segmentation in MMDA and MMDG tasks.  For MMDG, Li et al.~\cite{li2023bev} train models on two datasets and evaluate them on the remaining dataset. For MMDA, Jaritz et al.~\cite{jaritz2020xmuda}  identifies three adaptation scenarios: day-to-night adaptation (nuScenes Day/Night), country-to-country adaptation (nuScenes USA/Singapore), and dataset-to-dataset adaptation (A2D2/SemanticKITTI).
The Synthia~\cite{ros2016synthia} and Waymo~\cite{sun2020scalability} datasets are also widely  used in  MMTTA research. %For example, MM-TTA~\cite{shin2022mm} uses the Synthia-to-SemanticKITTI adaptation scenario and CoMAC~\cite{cao2023multi} identifies the SemanticKITTI-to-Synthia and SemanticKITTI-to-Waymo adaptation scenarios.
For the 2D semantic segmentation task, commonly used datasets include Cityscapes~\cite{cordts2016cityscapes}, GTA5~\cite{richter2016playing}, Synthia~\cite{ros2016synthia}, Mapillary~\cite{neuhold2017mapillary}, and ACDC~\cite{sakaridis2021acdc}.
Additionally, datasets such as UrbanSyn~\cite{gomez2023all}, Dark Zurich~\cite{sakaridis2019guided}, and BDD100K~\cite{yu2020bdd100k} are used in some studies to evaluate adaptation performance under varying environmental conditions. 

\subsection{Datasets for Image Classification}
Image classification is the most widely studied task for DA and DG with  in the context of multimodal foundation models. Among the most popular datasets are PACS~\cite{li2017deeper}, VLCS~\cite{fang2013unbiased}, Office-Home~\cite{venkateswara2017deep}, DomainNet~\cite{peng2019moment}, and Wilds~\cite{koh2021wilds} which serve as standard benchmarks for evaluating generalization across different domains. For adaptating multimodal foundation models, $11$ commonly used image classification datasets include ImageNet~\cite{deng2009imagenet}, Caltech101~\cite{fei2004learning}, OxfordPets~\cite{parkhi2012cats}, StanfordCars~\cite{krause20133d}, Flowers102~\cite{nilsback2008automated}, Food101~\cite{bossard2014food}, FGVCAircraft~\cite{maji2013fine}, SUN397~\cite{xiao2010sun}, DTD~\cite{cimpoi2014describing}, EuroSAT~\cite{helber2019eurosat} and UCF101~\cite{soomro2012ucf101}. Additionally, datasets such as ImageNetV2~\cite{recht2019imagenet}, ImageNet-Sketch~\cite{wang2019learning}, ImageNet-A~\cite{hendrycks2021natural}, and ImageNet-R~\cite{hendrycks2021many} are  widely used in  TTA setups to assess robustness under distribution shifts.

\subsection{Datasets for Other Applications}
Several benchmark datasets are widely used across different multimodal applications:

\begin{itemize}
    \item Multimodal sentiment analysis and emotion recognition tasks: CMU-MOSEI~\cite{zadeh2018multimodal}, IEMOCAP~\cite{busso2008iemocap}, MELD~\cite{poria2018meld}, MSP-IMPROV~\cite{busso2016msp}, and CMU-MOSI~\cite{zadeh2016mosi}.
    \item Multimodal person/vehicle re-identification: RGBNT201~\cite{zheng2021robust}, Market1501-MM~\cite{wang2022interact}, and RGBNT100 and RGBN300~\cite{li2020multi}. 
    \item Depth completion: VOID~\cite{wong2020unsupervised}, NYUv2~\cite{silberman2012indoor}, SceneNet~\cite{mccormac2016scenenet}, and ScanNet~\cite{dai2017scannet}. \item Domain generalized stereo matching: SceneFlow~\cite{mayer2016large}, ETH3D~\cite{schops2017multi}, Middlebury~\cite{scharstein2014high}, and KITTI 2015~\cite{menze2018object}.
    \item Cross-modal retrieval: COCO~\cite{chen2015microsoft}, COCO Narratives~\cite{pont2020connecting}, Open Narratives~\cite{pont2020connecting},  MSRVTT~\cite{pont2020connecting}, MSVD~\cite{chen2011collecting}, LSMDC~\cite{rohrbach2015dataset}.
    \item Medical image analysis: Vestibular schwannoma segmentation dataset~\cite{shapey2021segmentation}.
    \item Generative modeling: Storyboard20K~\cite{xie2024learning}.
    \item Face anti-spoofing: WMCA~\cite{george2019biometric}, CeFA~\cite{liu2021casia}, PADISI~\cite{rostami2021detection}, and SURF~\cite{zhang2020casia}. 
\end{itemize}

\section{Future Research Challenges}
\label{sec:future}
Despite significant progress, multimodal adaptation and generalization remain challenging and largely unsolved problems. In this section, we outline  key  future research directions, highlighting gaps in the current literature and discussing promising avenues for advancing the field.

\subsection{Theoretical Analysis}
While existing research has largely focused on the theoretical analysis of unimodal DA~\cite{ben2010theory} and DG~\cite{ye2021towards}, rigorous theoretical analyses in multimodal settings  are still lacking. Understanding the complexities of multimodal learning under domain shifts is crucial for developing more principled  MMDA and MMDG methods. SimMMDG~\cite{dong2023simmmdg} represents an early  in this direction, integrating multimodal representation learning with  domain generalization theory. Future work could benefit from bridging  multimodal learning theory~\cite{lu2023theory} with established unimodal DG principles to create  a unified theoretical framework for multimodal adaptation. Additionally, the theoretical analysis of adapting MFMs remain largely unexplored, leaving room for further investigation into their generalization properties, robustness under distribution shifts, and optimal adaptation strategies.

\subsection{Large-scale Benchmark and Datasets} 
Unlike unimodal DA and DG, which benefit from numerous well-established benchmarks~\cite{peng2019moment,koh2021wilds,zhang2023nico}, there are currently few benchmarks specifically designed for MMDA or MMDG. Additionally, the datasets used in existing MMDA and MMDG studies are significantly smaller in scale compared to their unimodal counterparts, limiting their ability to evaluate real-world generalization effectively. Future research should prioritize the  development of comprehensive multimodal  benchmarks encompassing datasets of varying scales, domains, and modalities. Establishing standardized evaluation protocols will be crucial for fostering fair comparisons and accelerating advancements in MMDA and MMDG methodologies.

\subsection{Open-set Settings} 
Most existing approaches operate under the closed-set assumption, which presumes identical label spaces across domains. However, in real-world applications, target domains often contain unknown classes, making it essential  to detect and handle them effectively. While  some recent studies~\cite{dong2024towards,dong2025aeo} have begun addressing the open-set scenario, this challenging yet practical setting remains underexplored.Further research is needed to develop robust multimodal open-set adaptation methods that can generalize across diverse domains while accurately identifying unseen categories. Additionally, techniques from  multimodal out-of-distribution detection~\cite{dong2024multiood,li2024dpu,liu2025fm} could  be leveraged and adapted to facilitate unknown class detection in multimodal adaptation and generalization.

\subsection{Research on MMTTA and MMDG} 
While considerable progress has been made in developing  MMDA methods, research on MMTTA~\cite{yang2023test,dong2025aeo} and MMDG~\cite{dong2023simmmdg,fan2024crossmodal} remains relatively limited. Given  the practical significance  of MMTTA and MMDG in real-world scenarios, future research should prioritize these areas.

\subsection{Diverse Downstream Tasks and Application Fields} 
Although multimodal adaptation and generalization have been extensively studied in tasks such as image classification,  action recognition, and semantic segmentation, their potential in other domains remains largely underexplored, including regression~\cite{cortes2011domain,nejjar2023dare}, generative models~\cite{yang2023one}, cross-modal retrieval~\cite{li2024test}, and image super-resolution~\cite{deng2023efficient}. Besides, the potential applications in other fields such as medicine, engineering, and cosmology remain underexplored and warrant greater attention.

\subsection{Missing Modality Robustness} 
Missing modality robustness remains a critical challenge in multimodal adaptation and generalization, particularly when certain  modalities (e.g., LiDAR, audio) are unavailable during deployment. While few existing approaches, such as cross-modal translation~\cite{dong2023simmmdg}, attempt to mitigate this issue, developing new frameworks that can dynamically prioritize available modalities and ensure robustness under missing-modality situations is crucial.

\subsection{MFMs for MMDA and MMDG} 
MFMs have hown significant promise in  enhancing unimodal DA and DG for images~\cite{fahes2023poda,cho2023promptstyler}. However, effectively leveraging MFMs to complement other modalities, such as audio and optical flow, remains an open challenge. Given their rich  feature representations and strong generalization capabilities, MFMs have the potential to further improve MMDA and MMDG performance.

% \subsection{Test-time Adaptation of VLMs} 
% Existing studies on VLM adaptation primarily employ supervised or few-shot learning setups that require labeled data. Test-time VLM adaptation~\cite{feng2023diverse,shutest,ma2024swapprompt}, which allows models to adapt dynamically during inference without labeled data, is a promising area for future exploration.
\subsection{Scalability and Efficiency} 
The deployment of large-scale MFMs is frequently constrained by their substantial computational demands, making efficiency a central research challenge. Reducing memory consumption and inference latency without sacrificing accuracy is crucial for real-world adoption. Techniques such as quantization~\cite{lee2021network}, pruning~\cite{liu2018rethinking}, and knowledge distillation~\cite{gou2021knowledge} offer promising directions. However, a core difficulty lies in compressing these models while preserving the fragile cross-modal alignments established during pre-training. Advances in lightweight architectures and optimization strategies will be essential to enable real-time multimodal inference on resource-constrained devices, ultimately facilitating the shift from cloud-only computation to practical edge deployment.

\subsection{The Problem of Negative Transfer} 
While numerous adaptation approaches for MFMs have demonstrated strong empirical performance, their limitations remain underexplored, particularly regarding failure modes and negative transfer. For instance, entropy minimization~\cite{shutest}, despite its popularity, can inadvertently amplify errors by reinforcing incorrect predictions under high uncertainty, resulting in overconfident misclassifications or even mode collapse. To advance the field, it is essential to prioritize robustness-oriented research. This entails developing reliable metrics to detect adaptation breakdowns, establishing standardized protocols for documenting instability, and fostering greater transparency by publishing negative findings and counterexamples. Such practices will not only expose systematic weaknesses but also guide the design of more resilient and trustworthy adaptation frameworks.

\subsection{Privacy Issues} 
Ensuring privacy and security is essential when adapting MFMs, especially in high-stakes areas such as healthcare and autonomous driving. The adaptation phase often involves handling sensitive or proprietary information, which increases the risk of unauthorized access or data leakage. At the same time, this stage leaves models vulnerable to adversarial attacks~\cite{madry2017towards} that can compromise updates, reduce reliability, and even cause harmful behavior. To address these risks, future research should prioritize privacy-preserving adaptation techniques. Federated learning~\cite{bao2023adaptive} represents a promising direction, enabling collaborative model improvement without direct exposure of raw data.

\subsection{Interpretability in Multimodal Adaptation} 
Most current multimodal adaptation methods operate as black boxes, making it difficult to understand their internal reasoning. A critical challenge is to develop methods that can explain how leveraging an auxiliary modality helps bridge the domain gap for a primary modality. For instance, how exactly does textual information assist a vision model in adapting to a new visual environment? Future work should focus on creating inherently interpretable models or post-hoc explanation techniques~\cite{samek2017explainable} tailored for multimodal adaptation. 
Moreover, the reasoning capabilities of MFMs~\cite{xia2025visionary} can be leveraged to enhance interpretability, for example, through chain-of-thought prompting, which allows models to expose intermediate reasoning steps rather than only providing final predictions.
Such methods would not only enhance model transparency and trustworthiness but also provide valuable insights for diagnosing failure modes and designing more effective cross-modal transfer strategies.

\subsection{Hybrid Architectures}
Future research should pursue hybrid multimodal architectures that integrate complementary modeling paradigms to overcome the inherent limitations of single-paradigm systems. CNN–Transformer hybrids, for instance, warrant particular attention: CNNs excel at efficient local feature extraction, while Transformers are powerful for capturing long-range dependencies and global context. Beyond two-stream hybrids, advanced ensemble strategies and Mixture-of-Experts (MoE)~\cite{shazeer2017outrageously} designs represent another promising direction. By dynamically routing modalities and domain-specific signals to specialized experts, MoE-based frameworks can improve scalability, efficiency, and robustness across heterogeneous datasets and deployment conditions. In addition, the development of unified hybrid architectures capable of addressing DA, DG, and TTA tasks within a single framework represents a promising research direction. This line of research holds strong potential for enabling adaptive, resource-aware multimodal models capable of meeting the demands of real-world applications.

\section{Conclusion}
\label{sec:con}
Adapting multimodal models to target domains under distribution shifts represents a critical challenge in machine learning that receives more and more attention these days. This survey provides a comprehensive overview of recent advancements in multimodal domain adaptation, multimodal test-time adaptation, and multimodal domain generalization, highlighting key challenges, methodologies, and applications driving progress in the field. Furthermore, we emphasize the critical role of multimodal foundation models in enhancing domain adaptation and generalization tasks, highlighting their potential to address real-world challenges across diverse modalities. By reviewing existing approaches, datasets, and applications, we identify several key directions for future research, including the development of better benchmarks and datasets, the handling of label shifts in dynamic environments, and further exploration of theoretical analysis. As the field continues to evolve, these insights offer a valuable foundation for advancing the robustness and efficiency of multimodal models in real-world scenarios and potentially transferring to other application fields such as medicine, engineering, and cosmology. %We hope this work inspires further advancements in this and related fields.

% use section* for acknowledgment
% \ifCLASSOPTIONcompsoc
% % The Computer Society usually uses the plural form
% \section*{Acknowledgments}
% \else
% % regular IEEE prefers the singular form
% \section*{Acknowledgment}
% \fi
% The authors acknowledge the support of "In-service diagnostics of the catenary/pantograph and wheelset axle systems through intelligent algorithms" (SENTINEL) project, supported by the ETH Mobility Initiative.

% Can use something like this to put references on a page
% by themselves when using endfloat and the captionsoff option.
\ifCLASSOPTIONcaptionsoff
\newpage
\fi

\bibliographystyle{IEEEtran}
\bibliography{my}

\vfill
	
\end{document}